\documentclass[10pt,twocolumn,letterpaper]{article}

\usepackage{iccv}
\usepackage{times}
\usepackage{epsfig}
\usepackage{graphicx}
\usepackage{amsmath}
\usepackage{amssymb}
\usepackage{mathrsfs}
\usepackage[scr=boondox]{mathalfa}

\DeclareGraphicsExtensions{.pdf,.png,.jpg}
\DeclareMathOperator*{\argmax}{argmax} 
\DeclareMathOperator*{\argmin}{argmin} 
\usepackage{makecell}
\usepackage{booktabs}
\usepackage{multirow}
\usepackage{here}
\usepackage{kotex}
\usepackage{stackengine}

\usepackage[breaklinks=true,bookmarks=false]{hyperref}

\iccvfinalcopy 

\def\iccvPaperID{686} 

\begin{document}
\title{Not Only Look But Observe: Variational Observation Model of Scene-Level 3D Multi-Object Understanding for Probabilistic SLAM}

\author{ 	
	Hyeonwoo Yu \\
	Automation and Systems Research Institute, Seoul National University\\
	{\tt\small bgus2000@snu.ac.kr}
}
\maketitle

\begin{abstract}
We present NOLBO, a variational observation model estimation for 3D multi-object from 2D single shot.
Previous probabilistic instance-level understandings mainly consider the single-object image, not single shot with multi-object; relations between objects and the entire scene are out of their focus.
The objectness of each observation also hardly join their model.
Therefore, we propose a method to approximate the Bayesian observation model of scene-level 3D multi-object understanding.
By exploiting variational auto-encoder (VAE), we estimate latent variables from the entire scene, which follow tractable distributions and concurrently imply 3D full shape and pose.
To perform object-oriented data association and probabilistic simultaneous localization and mapping (SLAM),
our observation models can easily be adopted to probabilistic inference by replacing object-oriented features with latent variables.

\end{abstract}

\section{Introduction}
Object-oriented features find various aspects such as semantic scene understanding, task planning and autonomous driving \cite{slam++}.
Real-time object detection and high-level feature estimation are inevitable in various applications such as object recognition, data association and object-oriented simultaneous localization and  mapping (SLAM).
Recently, a plethora of real-time multi-object detection methods have been developed beyond category classification for a single object image \cite{fasterRCNN,yolo9000,SSD}.
In some of these existing multi-object detection methods, however, the estimation results are still bound to the object categories and the location on the image (bounding box), and hardly concern other details.

For data association and object-oriented SLAM, it is better to exploit the various object representations such as complete 3D shape as well as categories.
However, reconstruction methods such as \cite{kinectfusion,elasticfusion} and \cite{SfM} are extremely challenging to perform in real-time on multi-object; these methods are conducted after scanning the various viewpoints of each object.
Therefore, estimation methods which disentangle representations like 3D shape or viewpoint orientation from a 2D image have been carried out \cite{dataDriven3Dvoxel,categorySpecificSLAM}.
With the emergence of deep learning, direct inference methods are also developed \cite{synthesizing3dshapes,image2mesh,3dgan,3drecgan}.
Furthermore, by using the multi-object detectors or their network structures, estimation methods for instance-level understanding of multi-object are also studied \cite{voxelNet_yolo3D,3D-RCNN,categorySpecificSLAM}.
\begin{figure}[t]
	\centering
	\includegraphics[scale=0.05]{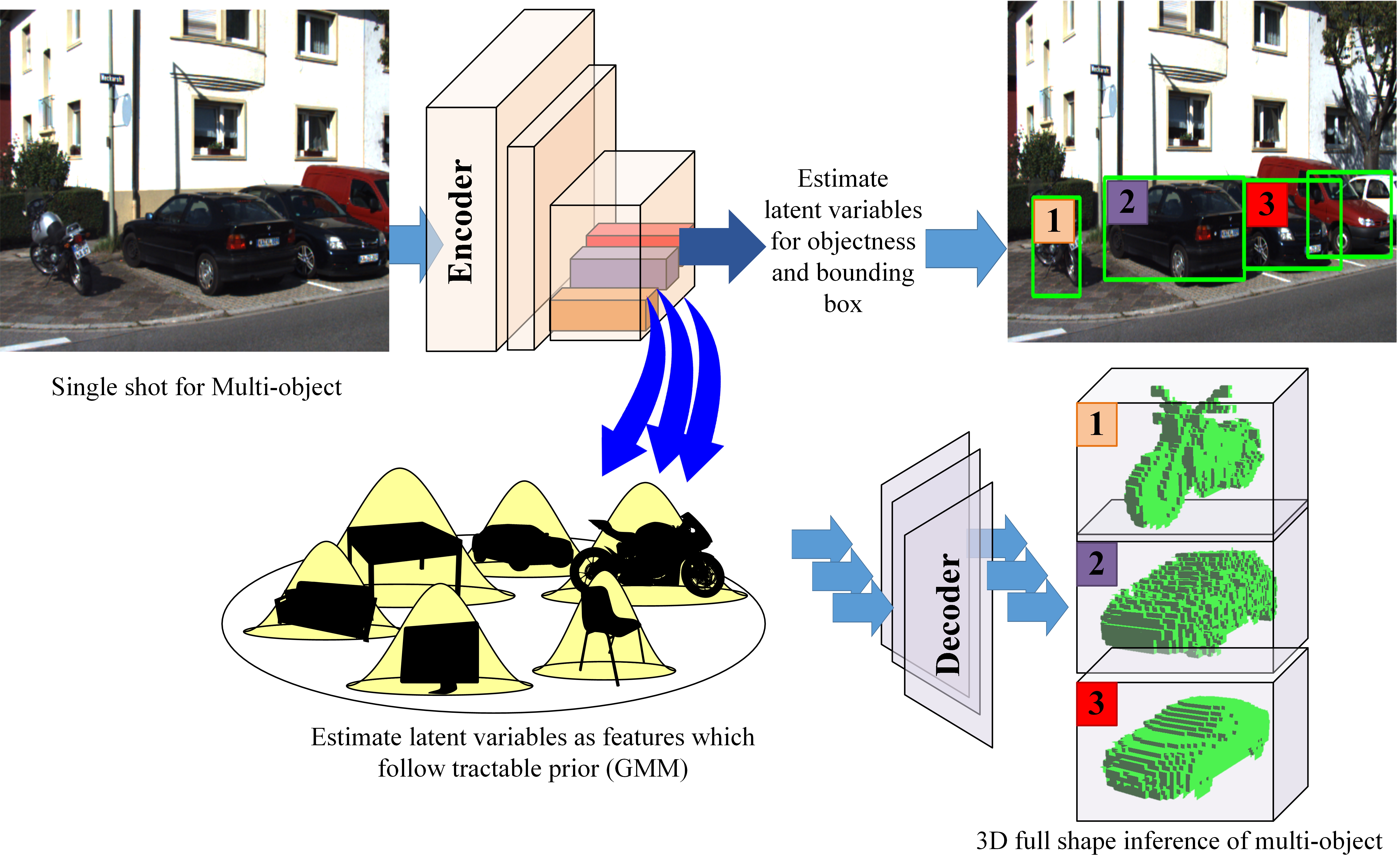}
	\caption{
		Overview of the proposed method.
		We train VAE to estimate the joint distribution of multi-object in single scene.
		Since objects are jointly related to each other and the scene, variational likelihoods for each observations are estimated from the entire scene simultaneously.
		Our model involves the object uncertainty, thus objectness can be reflect to the probabilistic SLAM.
		3D reconstruction also can be achieved by decoding the latent variables in parallel.
	}
	\label{likelihood_inference}
\end{figure}

These approaches, however, mainly focus on directly obtaining the disentangled features via network modeling, not probabilistic model.
Hence data association in Bayesian manner for probabilistic SLAM becomes challenging.
Even when outputs of hidden layers of the network are used as object features, it is still challenging to address the Bayesian inference since features follow intractable distributions \cite{modelnet,3dgan}.
As a result, in most cases there is little choice but to look at the outputs of well-designed networks;
they just look at the objects so far, but not observe them in the true sense.

In order to approximate the intractable object observation model, \cite{IROS2018} and \cite{ICRA2019} adopt the evidence lower bound (ELBO).
However, these methods mainly focus on a single object and hardly consider multi-object observation.
For single scene with multi-object, most of the instance-level understandings \cite{TLNet,3dgan,dataDriven3Dvoxel} and object-oriented SLAM \cite{ICRA2019,bowman2017probabilistic,categorySpecificSLAM,multimodalSLAM} should collect cropped region of interest (RoI) by using additional multi-object detector, and then input each image back into their models for single object.
Therefore, relations between objects, and between objects and single scene are out of their concern.
As object detector and observation model are separated from each other, objectness from multi-object detector is also hardly considered jointly with their work.


To this end, we propose a method for scene-level 3D multi-object understanding from single shot, and SLAM framework. 
We estimate the joint distribution of multi-object by using variational auto-encoder (VAE), considering the relations between objects and single scene.
The complex joint probability of multi-object can be captured in factorized form by leveraging the latent-space.
Latent variables are used instead of the object-oriented features for SLAM in Bayesian manner, thus frame-level understanding is easily reflected to SLAM formulations.
Since our model possesses latent variables for objectness, object uncertainty for each observation is also considered in probabilistic data association.
To achieve fast data association is crucial for real-time SLAM optimization, so we device a generative model that can reduce the dimensionality of latent variables.
Overview of our method is shown in Fig.~\ref{likelihood_inference}.

Our contributions are two-fold:
First, we mathematically show that the multi-object observation model considering relations between objects and single scene can be captured, by exploiting the existing multi-object detector structure.
Second, we introduce the probabilistic SLAM with our model, so that frame-level understanding and the object uncertainty are seamlessly reflected to the data association.



\section{Related work}

With the recent advent of neural networks, a number of single object classification and detection methods with high performance have been proposed \cite{imagenet,resnet,vggnet}.
Beyond obtaining one feature vector from one image for an object, several multi-object detection techniques from single shot have been developed by introducing new network structures \cite{fasterRCNN,yolo,yolo9000,SSD,retinanet}.
In particular, some of these methods can be applied to various real-time tasks since the whole detection network is composed of single network pipeline.

Various studies have also been conducted to understand the instance-level representation from 2D images such as object shape, orientation or bounding box.
\cite{renderforcnn,craftCNNforviewpoint,viewpointsandkeypoints} and \cite{3D-RCNN} estimate the orientation of the object by viewpoint classification with discretized bins.
In addition, 3D bounding box regression has been carried out to obtain the object location and orientation \cite{shuransong3Dbbox,seamlessSingleShot6D,6dofposefromsemantickeypoints,3dbboxdeeplearninggeo}. 
In order to estimate the distinct 3D shape of objects,
\cite{dataDriven3Dvoxel} aligns the prior shape to a single object image through key point matching and estimates its 3D shape and orientation together.
\cite{image2mesh} estimates the 3D mesh with linear combination of parameterized prior shapes.
In \cite{3dgan,TLNet,3drecgan,marrnet}, they have actively utilized non-linear regression and latent variables of neural networks for 3D reconstruction from 2D.

Through multi-object detection and instance-level understanding altogether, learning the disentangled representation of multi-object becomes achievable.
\cite{seamlessSingleShot6D} exploits the yolov2 structure \cite{yolo9000} to estimate the 3D bounding box and center of the multi-object and obtains the orientations.
In \cite{3D-RCNN}, they estimate the 3D shape rendering and orientation under faster R-CNN structure \cite{fasterRCNN}.
They obtain the shape rendering via weighted sum of the parameterized prior shape with PCL.
Orientations are estimated by classifying the bins which indicate the discretized object pose.
Similarly, in \cite{categorySpecificSLAM}, they design the object observation factor to perform data association for pose SLAM. 
RoIs for multi-object are obtained by \cite{yolo9000}.

These studies are efficient because they mainly concern direct and accurate estimation of the object characteristics through network modeling;
on the other hand, the probabilistic observation models are relatively less considered.
Although they exploit the neural network for nonlinear regression, approximating the intractable distribution is rarely concerned.
Therefore, Bayesian inference with obtained features are challenging;
for example, data association for SLAM is considered only in front-end and additional algorithms are necessary to perform loop closing and place recognition \cite{slam++,categorySpecificSLAM}.

To handle the intractable target distribution, latent variables can be adopted \cite{feibayesian,collapsedLDA,vae,gan}.
In order to understand and utilize the latent space, \cite{inversegraphics,disentanglingrepresentations} have studied the relations between latent variables and object visualization by using VAE \cite{vae}.
However, it is still challenging to apply the proposed method to probabilistic model approximation, as it mainly concentrates on the interpretable graphic codes.
To approximate the observation probability, entropy and variational likelihood is exploited in the field of the active vision \cite{informationtheoretic,towardsactiveevent,mlforactivevision}.
Using VAE, \cite{IROS2018,ICRA2019} have proposed methods to approximate the observation model of 3D objects for Bayesian inference.
Based on the ELBO which approximates the observation model, they have shown that how the probabilistic SLAM with data association can be performed with expectation-maximization (EM) algorithm.

However, the methods above only concern the observation model  for single object, thus it is inevitable to use multi-object detector for single-object images in the scene.
They hardly estimate the object observation model based on the entire scene. The relations between the objects and the scene are also merely considered.
It is also challenging to include the object uncertainty in their model, since the objectness is determined by the multi-object detector.
Therefore, we introduce the generative story of scene-level multi-object understanding for Bayesian inference, which is, to the best of our knowledge, the first of its kind.

\begin{figure*}[t]
	\centering
	\includegraphics[scale=0.21]{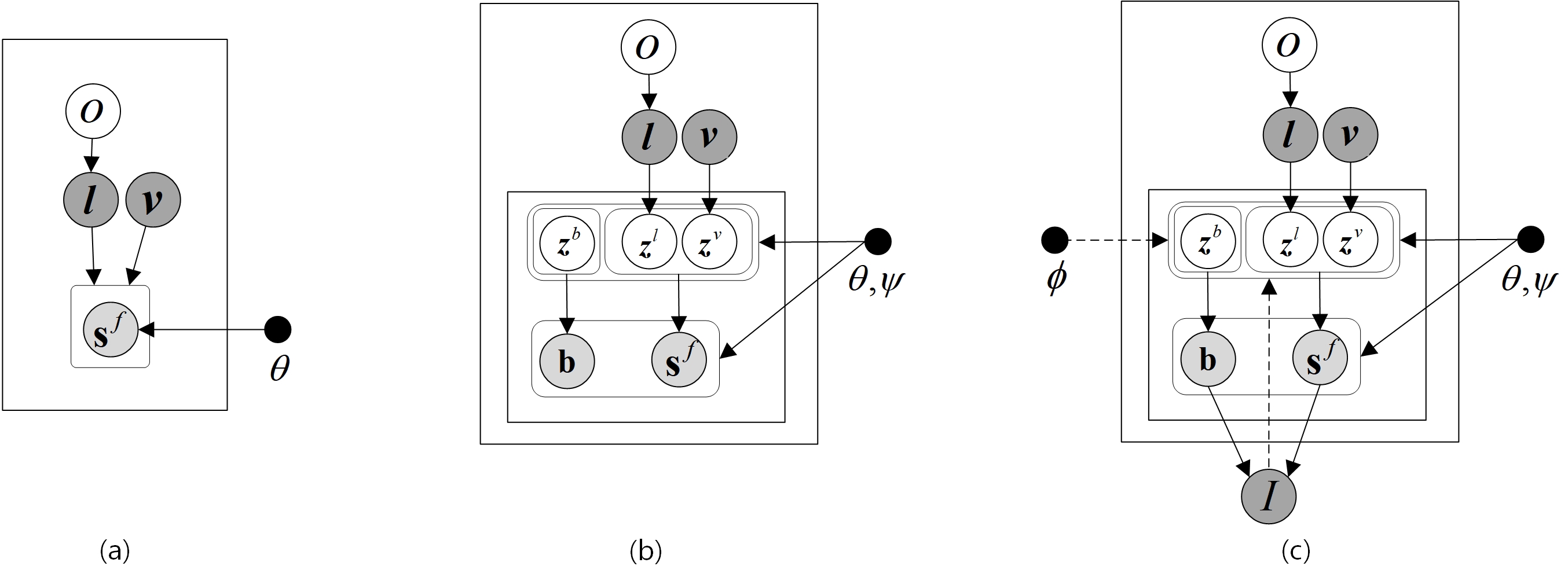}
	\caption{
		Overview of the proposed Bayesian graphical model for the object generative model.
		(a) The object label $\boldsymbol{l}$ and the orientation $\boldsymbol{v}$ related to the observer throw a Bayesian dice to generate a 3D full shape $\boldsymbol{s}^f$ of the object in the scene.
		For the objectness of an observed image, $o$, which generates $\boldsymbol{l}$, can be involved.
		(b) We exploit the latent variables to approximate the target distribution.
		Here, $\boldsymbol{z}^b$, $\boldsymbol{z}^l$ and $\boldsymbol{z}^v$ are for bounding box, basic 3D shape and orientation of the object, respectively.
		For the prior distributions of $\boldsymbol{z}^l$,  parameter $\psi$ is learned simultaneously with $\phi$ and $\theta$, which are the parameters of the encoder and the decoder, respectively.
		(c) We assume that single scene $I$ is generated by the 3D objects and their bounding boxes.
		Therefore, the variational likelihoods of each observation are estimated from $I$; relations between objects, and objects and single shot can be considered.
	}
	\label{graph_model}
\end{figure*}
\section{Multi-object Observation Model}
\subsection{Evidence Lower Bound and Encoder}
Suppose we select $K$ regions in a single scene and observe the full 3D shape of the arbitrary structure in each region.
The $k$'th area of the scene can be defined with RoIs \cite{fasterRCNN}, grids of fixed size \cite{yolo,yolo9000}, or grids of various sizes \cite{SSD,retinanet}.
The typical multi-object detection methods \cite{fasterRCNN,yolo9000,SSD,retinanet} mainly focus on the real-time detection and category inference.
For the generative story of object, however, any type of disentangled representations can be involved such as 3D shape or pose.
Let $\boldsymbol{s}^f_k \in \mathcal{S}^f$ be the $k$th observed \textit{f}ull 3D \textbf{\textit{s}}hape.
Similar to \cite{generatechair,ICRA2019}, we assume that the $\boldsymbol{l}$abel $\boldsymbol{l}_k \in \mathcal{L}$ and the $\boldsymbol{v}$iewpoint orientation $\boldsymbol{v}_k \in \mathcal{V}$ cast the Bayesian dice to generate the 3D shape as shown in Fig.~\ref{graph_model}(a):
$\boldsymbol{l}$ is the class or instance label of the 3D shape, and $v_i\in\boldsymbol{v}$ denotes the orientation $(SO3)$ of the shape related to the observer which can be represented in Euler angles.
To address the object location in the scene, $\boldsymbol{b}$ounding box $\boldsymbol{b}_k \in \mathcal{B}$ is also included in our story.

When observing single scene $I$mage $I$, objects are jointly related to each other since the objects and their locations determine $I$.
To catch the intractable observation model, the joint probability
$ p\left(
	\mathcal{S}^f, \mathcal{L}, \mathcal{V}, \mathcal{B}
\right)$ can be addressed.
Since our main concern is the object-oriented features, we solely focus on the 3D shape of the object excluding background in the scene.
For object and background discrimination, a latent variable $o_k\in\mathcal{O}$ for objectness can be added to the Bayesian graph model as shown in Fig.~\ref{graph_model}(a).
The joint probability
then can be fractionated as follows:
\begin{align}
\nonumber
	&\log \text{ } p\left(
		\mathcal{S}^f, \mathcal{L}, \mathcal{V}, \mathcal{B}
	\right)
	\\
	\nonumber
	&=
	\log
	\sum_{\mathcal{O}}
	p\left(\mathcal{L}, \mathcal{O}\right)
	p\left(\mathcal{V}\right)
	p\left(\mathcal{S}^f|\mathcal{L}, \mathcal{V}\right)
	p\left(\mathcal{B}\right)
	\\
	&=
	\log p\left(\mathcal{V}\right)
	+ \mathop{\underbrace{\log \sum_{\mathcal{O}}p\left(\mathcal{L}, \mathcal{O}\right)}}_{(a)}
	+ \mathop{\underbrace{\log p\left(\mathcal{S}^f | \mathcal{L}, \mathcal{V}\right)p\left(\mathcal{B}\right)}}_{(b)}
	\label{jointFrac}
\end{align} 
We let $\log p\left(\mathcal{V}\right) = c_0$ be a constant; that is, 3D shape can be observed in any arbitrary viewpoints.

To learn the complex probability distribution \eqref{jointFrac} using VAE, we first need to find the lower bound.
The joint probability $(a)$ in \eqref{jointFrac} can be denoted as $p\left(\mathcal{L}, \mathcal{O}\right) = p\left(\mathcal{L}\right) p\left(\mathcal{O}|\mathcal{L}\right)$.
Here, we assume that the prior $\log p\left(\mathcal{L}\right) = c_1$ is a uniform distribution, that is, any kind of structure can be detected in arbitrary location of a scene.
The lower bound of $(a)$ then can be represented as:
\begin{align}
\nonumber
	\log \sum_{\mathcal{O}} p\left(\mathcal{L}, \mathcal{O}\right)
	&=
	\log \sum_{\mathcal{O}} p\left(\mathcal{O}|\mathcal{L}\right) + c_1
	\\
	&\geq
	-KL\left(
		q_{\phi}\left(\mathcal{O}|I\right) || p\left(\mathcal{O}|\mathcal{L}\right)
	\right)
	+c_1.
	\label{olELBO|}
\end{align}
In our work, we assume that the entire scene $I$ is generated by objects and their locations;
therefore in \eqref{olELBO|} and the following, we let variational likelihood $q$ be estimated from $I$ in order to consider the correlations between objects and the entire scene.

Similarly, we can have ELBO of $(b)$ in \eqref{jointFrac} as the following:
\begin{align}
	\nonumber
	& \log p\left(\mathcal{S}^f|\mathcal{L}, \mathcal{V}\right)p\left(\mathcal{B}\right)
	\\
	\nonumber
	&\geq
	-KL\left(q_{\phi}\left(\mathcal{Z}^l,\mathcal{Z}^v|I\right)||p_\psi\left(\mathcal{Z}^l,\mathcal{Z}^v|\mathcal{L},\mathcal{V}\right)\right)
	\\
	\nonumber
	&+
	\mathop{\mathbb{E}}_{q}
	\left[
	\log p_{\theta}\left(\mathcal{S}^f|\mathcal{Z}^l,\mathcal{Z}^v\right)
	\right]
	\\
	&
	-KL\left(
	q_{\phi}\left(\mathcal{Z}^b|I\right) || p\left(\mathcal{Z}^b\right)
	\right)	
	+
	\mathop{\mathbb{E}}_{q}
	\left[
		\log p_{\theta}\left(\mathcal{B}|\mathcal{Z}^b\right)
	\right]
	\label{slvELBO}
\end{align}
The graph model including latent variable $\mathcal{Z}=\mathcal{Z}^l,\mathcal{Z}^v,\mathcal{Z}^b$ in \eqref{slvELBO} is shown in Fig.~\ref{graph_model}(b).
With \eqref{olELBO|} and \eqref{slvELBO}, the lower bound of \eqref{jointFrac} can be achieved; however, the formulation is composed of the joint probability of latent variables for entire objects, which is still intractable.

To relax the problem, we can have the lower bound in factorized form for each object by adopting the mean field inference \cite{vae}.
We assume that all elements of latent variables $\mathcal{O}$ and $\mathcal{Z}$ are independent to each other. That is, the lower bound can be factorized with $o_k, \boldsymbol{z}_k\in\mathcal{O},\mathcal{Z}$ which are for $k$'th region in $I$.
The lower bound $L$ of \eqref{jointFrac} then
can be represented as:
\begin{align}
	\nonumber
	&L
	=
	\sum_{\boldsymbol{l}_k\in \mathcal{L}^o \cup \mathcal{L}^{\neg o}}
			-KL\left(
			q_{\phi}\left(o_k|I\right) || p\left(o|\boldsymbol{l}_k\right)
			\right)
	+
	\\
	\nonumber
	&
	\sum_{\boldsymbol{l}_k\in\mathcal{L}^o}
	\big(
			-KL(\mathop{q}_{\phi}\left(\boldsymbol{z}^{lv}_k|I\right)
			||
			\mathop{p}_\psi\left(\boldsymbol{z}^{lv}_k|\boldsymbol{l}_k,\boldsymbol{v}_k\right)
			)
	+
	\mathop{\mathbb{E}}_{q}
	[
		\log
		\mathop{p}_{\theta}(
			\boldsymbol{s}^f_k | \boldsymbol{z}^{lv}_k
		)
	]
	\big)
	\\
	&+
	\sum_{\boldsymbol{l}_k\in\mathcal{L}^o}
	\big(
			-KL\left(q_{\phi}\left(\boldsymbol{z}^b_k|I\right)
			||
			p_\psi\left(\boldsymbol{z}^b_k\right)
			\right)
	+
			\mathop{\mathbb{E}}_{q}
			\left[
			\log
			{p}_{\theta}\left(
				\boldsymbol{b}_k | \boldsymbol{z}^b_k
			\right)
			\right]
	\big)
	\label{summed_lowerbound}
\end{align}
where $\boldsymbol{z}^{lv} = \boldsymbol{z}^l, \boldsymbol{z}^v$. $\mathcal{L}^o$ and $\mathcal{L}^{\neg o}$ are the object and non-object (background) label set respectively.
In \eqref{summed_lowerbound}, the first row can be viewed as the lower bound for objectness, second for 3D shape reconstruction and third for bounding box regression. 
The KL and expectation terms can be learned using encoding and decoding parts of VAE, respectively.

In this manner, joint observation model for multi-object can be captured as factorized form in latent space. Each $\boldsymbol{z}_k$ is for each observation but estimated from the entire scene $I$, and relations between objects are naturally taken into account.
To implement the network for each observation, $K$ encoders are required.
However, since the variational likelihoods of each observation share the parameter $\phi$ and are estimated from $I$ in common, encoders can be combined in single encoder which estimates all likelihoods simultaneously. In our work, we exploit YOLOv2-like structure as an encoder, which enables the real-time performance and end-to-end learning scheme.
The graphical model with variational likelihood estimation is depicted in Fig.~\ref{graph_model}(c).

\subsection{Low-Dimensional Latent Variables and Decoder}
In order to learn the posterior $p_{\theta} \left(\boldsymbol{s}^f | \boldsymbol{z}^{lv}\right)$ for 3D shape reconstruction in \eqref{summed_lowerbound}, we can use 3D decoder which outputs the rotated 3D shape according to the observed viewpoint.
When the algorithmic prior operation such as rotation transform exists, however, separating such arithmetically trivial operation from the non-linear regression technique can relieve the whole network, and enable the efficient learning \cite{carneiro2010fusion,krishnan2015drought,jonschkowski2018differentiable,qian2018deep}.
For the shape reconstruction term of \eqref{summed_lowerbound}, we can let $\boldsymbol{s}^f = f_R\left(\boldsymbol{s}^{fo}, \boldsymbol{R}\right)$, where $f_R$ is a function that rotates the basic orientation shape $\boldsymbol{s}^{fo}$ with rotation matrix $\boldsymbol{R}$ arithmetically. Then we have:
\begin{align}
	\nonumber
	\mathop{\mathbb{E}}_{q}
	\left[
		p_{\theta}\left(\boldsymbol{s}^f|\boldsymbol{z}^{lv}\right)
	\right]
	&=
	\mathop{\mathbb{E}}_{q}
	\left[
		p_{\theta}\left(\boldsymbol{s}^{fo}, \boldsymbol{R}^f | \boldsymbol{z}^{lv}\right)
	\right]
	\nonumber
	\\
	&=
	\mathop{\mathbb{E}}_{q}
	\left[
		p_{\theta}\left(\boldsymbol{s}^{fo}|\boldsymbol{z}^l\right)
	\right]
	\mathop{\mathbb{E}}_{q}
	\left[
		p\left(\boldsymbol{R}^f|\boldsymbol{z}^v\right)
	\right]
	\nonumber
	\\
	&=
	\mathop{\underbrace{
		\mathop{\mathbb{E}}_{q}
		\left[
			p_{\theta}\left(\boldsymbol{s}^{fo}|\boldsymbol{z}^l\right)
		\right]
	}}_{(a)}
	\mathop{\underbrace{
		\mathop{\mathbb{E}}_{q}
		\left[
			p\left(\boldsymbol{R}^f|\boldsymbol{R}^z\right)
		\right]
	}}_{(b)}
	\label{shapeDecoder}
\end{align}
where $\boldsymbol{R}^f$ is the rotation matrix according to the shape $\boldsymbol{s}^f$.
$\boldsymbol{R}^z$ is the rotation matrix computed from $\boldsymbol{z}^v$, and thus we let $\boldsymbol{z}^v$ be the trigonometric value of Euler angles as in \cite{ICRA2019}.
Since we choose the binary voxelized grid to represent the 3D shape, $(a)$ in \eqref{shapeDecoder} is assumed to be binary distribution. For orientation, we let $(b)$ follow a Von Mises–Fisher distribution \cite{directional,directional1}.

With \eqref{shapeDecoder}, the second row of \eqref{summed_lowerbound} can be expressed as the following:
\begin{align}
	\nonumber
	&-KL\left(
	{q}_{\phi}\left(\boldsymbol{z}^{lv}|I\right)
	||
	{p}_\psi\left(
			\boldsymbol{z}^{lv}|\boldsymbol{l},\boldsymbol{v}
		\right)
	\right)
	+
	\mathop{\mathbb{E}}_{q}
	\left[
		\log
		{p}_{\theta}\left(
			\boldsymbol{s}^f | \boldsymbol{z}^{lv}
		\right)
	\right]
	\\
	\nonumber
	&=
	-KL\left(
		q_\phi\left(
			\boldsymbol{z}^l | I
		\right)
		||
		p_\psi\left(
			\boldsymbol{z}^l | \boldsymbol{l}
		\right)
	\right)
	+
	\mathop{\mathbb{E}}_{q}
	\left[
		\log
		p_\theta\left(
			\boldsymbol{s}^{fo} | \boldsymbol{z}^l
		\right)
	\right]
	\\
	&-
	KL\left(
		q_\phi\left(
			\boldsymbol{z}^v | I
		\right)
		||
		p\left(\boldsymbol{z}^v | \boldsymbol{v}\right)
	\right)
	+
	\mathop{\mathbb{E}}_{q}
	\left[
		p\left(
			\boldsymbol{R}^f | \boldsymbol{R}^z
		\right)
	\right]
	\label{decoder_loss}
\end{align}
For the tractable prior distribution of $\boldsymbol{z}^l$, we assume $p_\psi\left(\boldsymbol{z}^l|\boldsymbol{l}\right) = \mathcal{N}\left(\boldsymbol{z}^l;\boldsymbol{\mu}\left(\boldsymbol{l}\right), \boldsymbol{I}\right)$.
In other words, we let $p_\psi\left(\boldsymbol{z}^l\right) = \sum_{\boldsymbol{l}} p\left(\boldsymbol{z}|\boldsymbol{l}\right) p\left(\boldsymbol{l}\right)$ be a gaussian mixture model (GMM) as in \cite{imagedescriptionGMMVAE,ICRA2019}.
$\boldsymbol{\mu}\left(\boldsymbol{l}\right) = f_\psi \left(\boldsymbol{l}\right)$ is obtained from prior network with parameter $\psi$, which is trained with VAE simultaneously.
The variational likelihoods except for $q_\phi\left(o|I\right)$ are assumed to be isotropic Gaussians. For more details of each distribution, see Appendix I.

Our decoder estimates only the basic shape without considering the orientation relative to the observer. To complete observed shape inference, rotation transform on $\boldsymbol{R}^f$ should be performed subsequently. In this way, we can relieve the burden of our network and reduce the parameters of it; we empirically found that the dimension of the latent variable $\boldsymbol{z}^l$ can be decreased from $128$ to $16$, which is crucial for SLAM performance.
As described in the next section, latent variables can replace the object-oriented features, and the metric operation between each feature is inevitable for data association. Reduced dimensional latent variables thus make the SLAM optimization process up to 5 times faster. Additionally, with the low-dimensional Gaussian latent variables, the bubble effect is relaxed \cite{hypersphericalVAE} and $p_\psi\left(\boldsymbol{z}^l|\boldsymbol{l}\right)$ becomes distinct according to the label $\boldsymbol{l}$.
Hence, robust data association is achieved.


\section{Latent Variables and Probabilistic SLAM}
Consider the localization and mapping problem with object-oriented features.
Let $\mathcal{X}=\{\mathscr{x}_t=\left(\boldsymbol{x}^{\mathscr{x}}_t, \boldsymbol{v}^{\mathscr{x}}_t\right)\}^T_{t=1}$ be the pose of an observer.
Assume we have a collection $\mathscr{L}=\{\ell_m = \left(\boldsymbol{l}_m, \boldsymbol{x}^{\ell}_m, \boldsymbol{v}^{\ell,global}_m\right)\}_{m=1}^M$ of $M$ landmarks.

Now suppose that the observer navigates around the area and obtain a set of observation $\mathcal{S}=\{\mathcal{S}^T_{t=1}\}$ for $T$ keyframes.
Here, $k$'th observation $\boldsymbol{s}_k = \left(\boldsymbol{s}^f_k,\boldsymbol{x}^{\boldsymbol{s}}_k\right) \in \mathcal{S}_t$
can be about either an object or background.
In the previous works for object-oriented SLAM \cite{bowman2017probabilistic,categorySpecificSLAM,ICRA2019,multimodalSLAM}, they lean to the existing multi-object detection method to get rid of the non-object detection; frame-level joint probability and objectness are hardly considered when formulating SLAM optimization.
In our method, since objectness joins our single scene understanding and thus naturally affects the data associations, all observations obtained from the regions of $I$ can be seamlessly used for SLAM.

Adopting the lower bound \eqref{summed_lowerbound} as approximated observation model, the optimal $\mathcal{X}$ and $\mathscr{L}$ for the probabilistic SLAM is obtained from the maximization step of EM formulation:
\begin{align}
\nonumber
&\mathcal{X}, \mathcal{X}^\ell
\\
\nonumber
&=
\argmin_{\mathcal{X}, \mathcal{X}^\ell}
\sum_{t,k,j}
-w^t_{kj}
\log
p\left(
\boldsymbol{x}^{\boldsymbol{s}}_k | \boldsymbol{x}^\mathscr{x}_t,\boldsymbol{x}^\ell_j
\right)
p \left( \boldsymbol{z} = \boldsymbol{v}_k  | \boldsymbol{v}^{l,local}_{t,j}\right),
\\
&\mathcal{L}^M
=
\argmin_{\mathcal{L}^M}
\sum_{t,k,j}
-w^t_{kj}
\log
p_\psi \left( \boldsymbol{z} = \boldsymbol{\mu}^{\boldsymbol{s}l}_k | \boldsymbol{l}_{j}\right)
\label{EM_maximization_frac_label}
\end{align}
where $\mathcal{X}^\ell = \{\boldsymbol{x}^\ell_m, \boldsymbol{v}^{\ell,global}_m\}_{m=1}^M$ and $\mathcal{L}^M=\{\boldsymbol{l}_m\}_{m=1}^M$.

Before calculating \eqref{EM_maximization_frac_label}, in expectation step, the similarity weight $w^t$ is calculated in consideration of the objectness $\mathcal{O}$ of observations.
Note that the object-oriented feature is replaced with encoded variables $\boldsymbol{v}_k$ and $\boldsymbol{\mu}^{\boldsymbol{s}l}_k$ for each $k$'th grid, even though we start with joint probability of $t$'th keyframe.
Also, for the likelihood of object-oriented feature, tractable latent priors $p\left(\boldsymbol{z}|\boldsymbol{v}\right)$ and $p_\psi$ are used which are isotropic Gaussians.
Therefore, with simple derivations, optimal solutions can be achieved even if the inaccurate observations are made, as the objectness of the multiple observation is concurrently considered.
Details of the EM formulation can be found in Appendix II.

\begin{figure*}[t]
	\centering
	\includegraphics[scale=0.11]{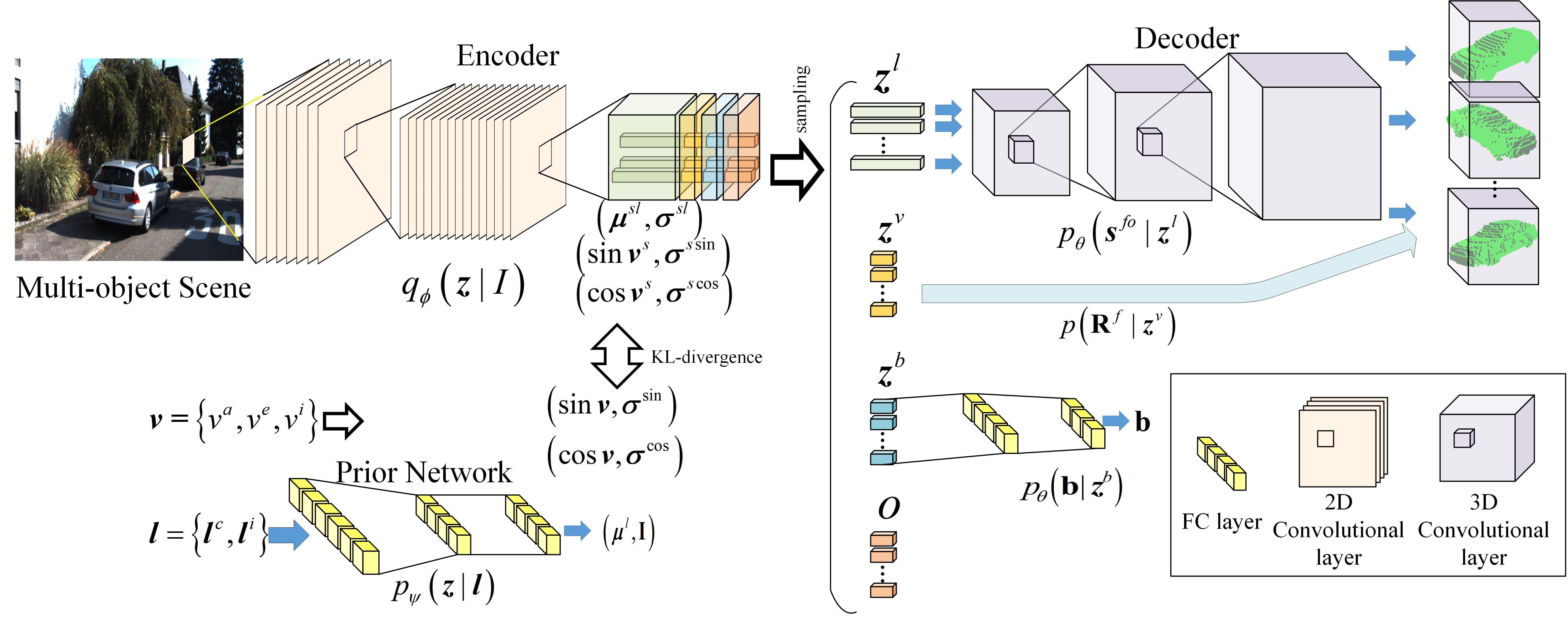}
	\caption{
		Proposed network architecture for variational 3D object observation.
		We use darknet-19 in YOLOv2 as a core network of the encoder. The generator in 3D-GAN is adopted for the decoder.
		We set the dimension of the latent variable to $16$, since the rotation task is separated to the decoder.
		For the prior network consists of fully connected (FC) layers, and is trained with the encoder and the decoder simultaneously.
		When the encoder network has single scene and infers latent variables for each RoI, we only select the latent variables of objects; therefore, the decoder gets as many latent variables as objects in a scene.
		In other words, the batch size varies between the encoder and the decoder when training as well as test.
		Our network can be trained in an end-to-end manner.
	}
	\label{proposed_networks}
\end{figure*}
\section{Implementation}
To implement the proposed observation network, in this paper we use darknet19 structure \cite{yolo9000} for the encoder core (or backborn).
We construct the encoder by adding 3 convolutional layers with 1024 filters followed by one convolutional layer on top of the core network.
A predictor (encoder) predicts $q_\phi$ for each grid.
In other words, predictor infers $q_\phi\left(o|I\right)$ for objectness, $q_{\phi}\left(\boldsymbol{z}^l|I\right)$ for latent variables implying full shape, $q_{\phi}\left(\boldsymbol{z}^v|I\right)$ for viewpoint inference, and
$q_\phi\left(\boldsymbol{z}^b|I\right)$ for bounding box.
The decoder follows the generator structure of \cite{3dgan}, except the input dimension; which in our case is set to 16.
A prior network consists of dense layers to represent GMM for prior distribution $p_\psi$.
As in \cite{ICRA2019}, prior network is trained with VAE simultaneously.

Similar to \cite{yolo9000}, when training we consider one predictor which predicts the highest IOU with the ground truth bounding box as responsible predictor for an object.
After selecting the predictors observing the object in the grids, shape estimation is performed by inputting the latent variables obtained from that predictors to the decoder.
Therefore, during both training and testing, the input batch size varies between the encoder and decoder; when single scene enters the encoder, the decoder gets as many latent variables as the number of objects in the scene.
The proposed network structure is displayed in Fig.~\ref{proposed_networks}.

\section{Training details}
The proposed network estimates the various representations of multi-object in single scene, with probabilistic distributions.
The negative lower bound from \eqref{summed_lowerbound}, which is composed of KL divergence and expectation terms of various distributions, is used as the training loss. The network thus easily diverged without sophisticated training strategy.
For the stable optimization procedure, we replace the objectness term in \eqref{summed_lowerbound} with $KL\left(
p\left(o|\boldsymbol{l}\right)
||
q_{\phi}\left(o|I\right)
\right)$ in actual training.
Then objectness loss becomes equal to the conventional binary cross-entropy loss.

We also found that the two-stage pretraining make the main training stabilized: pre-pretraining of 2D-3D understanding, and pretraining of NOLBO for single object.
We first pre-pretrain the encoder core on the ImageNet dataset \cite{imagenetDataset} for object classification and Render for CNN dataset \cite{renderforcnn} for viewpoint classification sequentially.
In the case of the decoder, it can fall into the local minimum when learning a small number of object instances, since the decoder only infers the basic 3D shape without considering the rotation transform.
To alleviate this limitation, the decoder is also pre-pretrained on ModelNet40 dataset \cite{modelnet} which contains 40 classes, and about 300 instances per each class.
We construct 3D VAE with decoder, 3D encoder in \cite{3dgan} and prior network for this pre-pretraining.


The NOLBO approximates the observation model by learning 3D reconstruction and pose estimation for multi-objects in 2D scenes.
Therefore, we use Pascal3D+ \cite{pascal3D} and objectnet3D \cite{objectnet3D} training datasets comprising 2D-3D aligned annotations considering 3D shape pose.
Object orientation $\boldsymbol{v}$ is expressed as azimuth, elevation and in-plane rotation angles.
Since these datasets contain 100 classes in total, we manually select 40 classes.
Prior to the NOLBO multi-object training, we pretrain NOLBO for single-object \cite{IROS2018,ICRA2019} on these datasets with pre-pretrained encoder core, decoder and a fresh prior network.
As \cite{yolo9000}, networks are trained on multi-scale images.
Since the networks slowly converge when trained on multi-resolution images from the beginning, we first fix the image resolution to $224\times224$, and change to $448\times448$ when the accuracy of 3D reconstruction reaches about 70\% mAP.
Onece the accuracy reaches 70\% again, multi-scale training is started.

The networks for NOLBO multi-object are constructed based on encoder core, decoder and prior network of the pretrained NOLBO single-object.
Training starts with the multi-resolution images.
We use adam optimizer with a starting learning rate of $10^{-5}$ for the first epoch, and increase to $10^{-4}$.
For the 20 epochs, we freeze decoder and prior network as they already learn the shape distribution.
This allows the network to learn how to infer the 3D shape distribution from 2D scene without diverging.
Similar to \cite{yolo9000,ICRA2019}, gaussian blur, HSV saturation, RGB invertion and random brightness are applied to 2D scene data augmentation.
Random translation and scaling are also used.

All of our code and pre-trained models are available at \url{https://github.com/bogus2000/NOLBO}.

\begin{figure*}[t]
	\centering
	\includegraphics[scale=0.50]{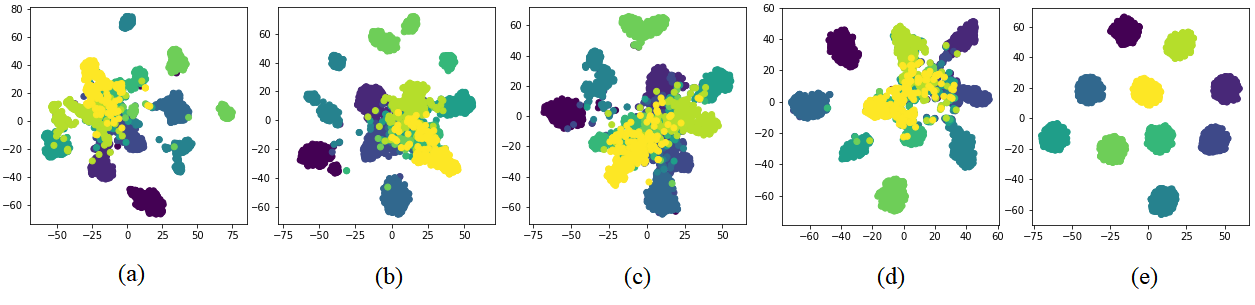}
	\caption{
		Comparison of the latent spaces of 
		(a) TLNet, (b) 2D-3D auto-encoder, (c) vanilla VAE and (d) NOLBO.
		We also plot the latent space of the prior distribution $p_\psi\left(\boldsymbol{z}|\boldsymbol{l}\right)$ in (e), which is trained simultaneously.
		To visualize the prior distribution, we sample the latent variables from $p_\psi\left(\boldsymbol{z}|\boldsymbol{l}\right)$ and plot them.
		Latent variables are colorized according to their respective object categories.
		Some of the latent variables in (a-c) are separately distributed even in the same categories.
		In the case of NOLBO, latent variables tend to be grouping according to their categories, as the KL-divergence terms in \eqref{summed_lowerbound} enforce the latent variables to follow the prior distribution displayed in (e).
	}
	\label{latentdistribution}
\end{figure*}

\begin{table*}[h]	
	\begin{center}
		\begin{tabular}{c | c c c c c c c c c c c c | c}
			\hline
			& aero & bike & boat & bottle & bus & car & chair & table & mbike & sofa & train & tv & mean \\
			\hline
			\textit{Acc}$_{\frac{\pi}{6}}$ (\cite{viewpointsandkeypoints}) & 0.81 & 0.77 & 0.59 & \textbf{0.93} & \textbf{0.98} & 0.89 & 0.80 & 0.62 & \textbf{0.88} & 0.82 & 0.80 & 0.80 & 0.81 \\
			\textit{Acc}$_{\frac{\pi}{6}}$ (\cite{3dbboxdeeplearninggeo}) & 0.78 & 0.83 & 0.57 & \textbf{0.93} & 0.94 & 0.90 & 0.80 & 0.68 & 0.86 & 0.82 & 0.82 & 0.85 & 0.81 \\
			\textit{Acc}$_{\frac{\pi}{6}}$ (Ours) & \textbf{0.83} & \textbf{0.86} & \textbf{0.70} & 0.90 & 0.95 & \textbf{0.96} & \textbf{0.95} & \textbf{0.83} & 0.83 & \textbf{0.98} & \textbf{0.94} & \textbf{0.91} & \textbf{0.88} \\
			\hline
			\textit{MedErr} (\cite{viewpointsandkeypoints}) & 13.8 & 17.7 & 21.3 & 12.9 & 5.8 & 9.1 & 14.8 & 15.2 & 14.7 & 13.7 & 8.7 & 15.4 & 13.6 \\
			\textit{MedErr} (\cite{3dbboxdeeplearninggeo}) & \textbf{13.6} & \textbf{12.5} & 22.8 & \textbf{8.3} & \textbf{3.1} & \textbf{5.8} & 11.9 & \textbf{12.5} & \textbf{12.3} & 12.8 & \textbf{6.3} & \textbf{11.9} & \textbf{11.1} \\
			\textit{MedErr} (Ours) & 14.5 & 16.5 & \textbf{17.8} & 10.5 & 10.1 & 8.6 & \textbf{11.4} & 13.7 & 16.9 & \textbf{10.7} & 9.2 & 14.1 & 11.7 \\
			\hline
		\end{tabular}
	\end{center}
	\caption{Comparison of the Viewpoint Estimations with Ground Truth Bounding Box on Pascal3D+ test dataset}
	\label{viewpointestimation}
\end{table*}

\begin{figure}[h]
	\centering
	\includegraphics[scale=0.28]{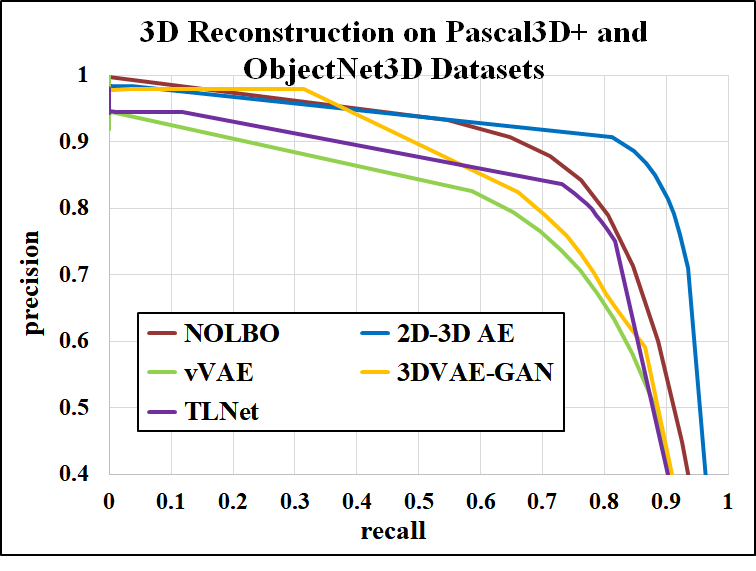}
	\caption{
		Precision-recall curve of 3D reconstructions on Pascal3D+ and Objectnet3D test dataset.
	}
	\label{precision_recall}
\end{figure}
\begin{figure*}[t]
	\centering
	\includegraphics[scale=0.18]{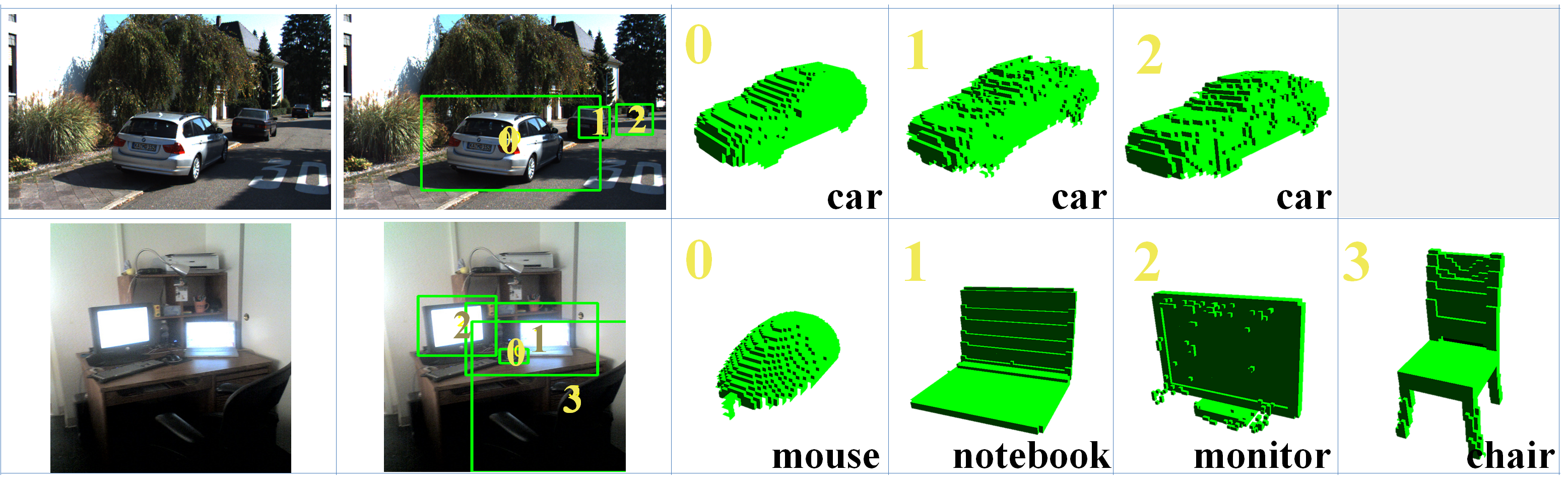}
	\caption{
		Examples of the 3D shape estimations of multi-object and classifications using MLE.
		We display several reconstruction results of objects in each 2D scene.
		As shown in the first row, the instance-level 3D shape estimation is achievable.
	}
	\label{multiobject3Dreconstruction}
\end{figure*}
\begin{figure*}[t]
	\centering
	\includegraphics[scale=0.18]{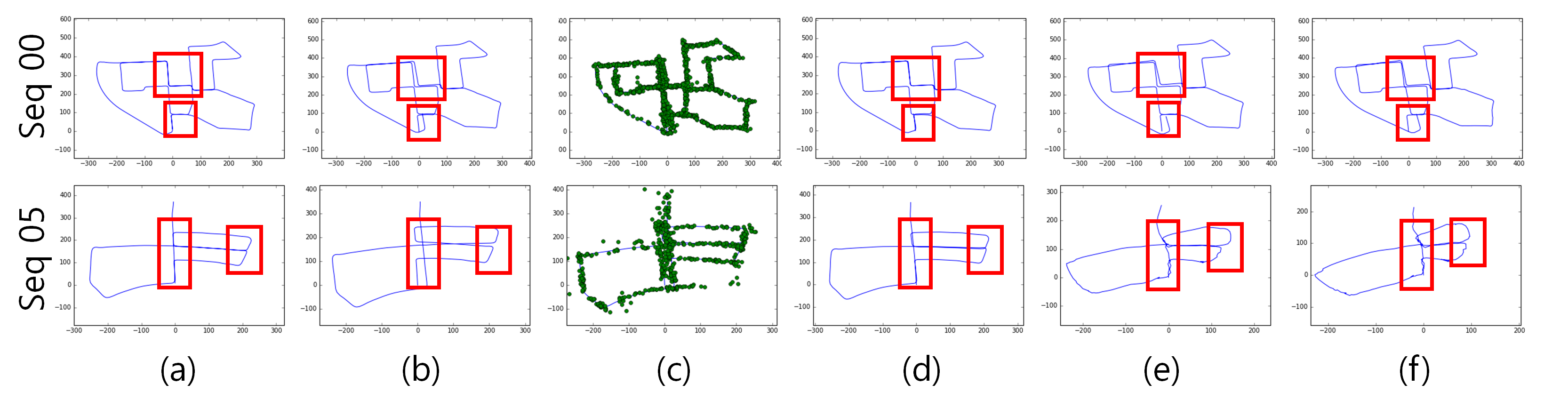}
	\caption{
		Several trajectory estimation results. we mark the important loop closing regions with red boxes. (a) Ground truth. (b) Visual odometry with known scale. (c) Observed objects. SLAM results using (d) NOLBOMulti, (e) NOLBOSingle and (f) vanilla VAE.
	}
	\label{trj}
\end{figure*}

\section{Experiments}
We evaluate the proposed method in various aspects; disentangled representations and SLAM application.
The main purpose of NOLBO is to approximate the object observation model.
When the model is applied to the Bayesian inference, latent variables and their prior distributions are used as object-oriented features and their observation models.
Therefore, it is important how objects are located or projected on the latent space as latent variables.
For comparison of latent spaces, we construct a 2D-3D auto-encoder (AE), vanilla VAE (vVAE) \cite{vae} and TL-Network (TLNet) \cite{TLNet} by using our pretrained encoder and decoder structure.
We also compare the SLAM results using object-oriented features from above networks.

\subsection{Object Pose Estimation}
Since NOLBO approximates the observation model, it is possible to perform category classification and viewpoint estimation using MLE (see APPENDIX II).
The latent variables for orientation becomes the orientation itself, so viewpoint orientation can directly be inferred from the encoder.
To report the qualitative results of our pose estimation method, we train NOLBO with 2D images of fixed size.
The comparison of the viewpoint estimation is shown in Table \ref{viewpointestimation}.
Our method shows competitive results relative to previous works.

\subsection{Object Observation and Latent Space}
For the data association or MLE, NOLBO have advantages that latent variables follow distinct prior distributions for each class or instance.
In order to display how objects are projected into the latent space according to their respective class, we estimate the latent variables from NOLBO and other methods.
We use the outputs from 2D encoders of other non-probabilistic methods as latent variables.
The dimensions of the latent variables are set to 16, same as NOLBO.
We display the latent variables obtained from each network in Fig.~\ref{latentdistribution}.
We use t-sne method \cite{tsne} for the latent space dimension reduction in order to plot in 2D.
For clarity, we randomly choose 10 classes and show the results.
Since NOLBO enforce the latent variables to follow the prior distributions which are learned together, the latent variables are projected to particular distributions according to their categories unlike other methods.

\begin{table*}[h!]	
	\begin{center}
		\begin{tabular}{c | c | c | c | c | c | c | c }
			\hline
			& NOLBOMulti16 & NOLBOMulti128 & NOLBOSingle16 & AE16 & TLNet16 & vVAE16 & VO \\
			\hline
			ATE & 25.93 & 28.66 & 52.79 & 37.88 & 62.56 & 64.78 & 30.75 \\
			\hline
			RPE & 29.88 & 31.71 & 65.03  & 40.83 & 73.02 & 72.07 & 35.40 \\
			\hline
		\end{tabular}
	\end{center}
	\caption{Average precisions of the trajectory estimation results evaluated on KITTI dataset}
	\label{ATEandRPE}
\end{table*}
\subsection{Object 3D Shape Reconstruction}
For comparison of the 3D reconstruction results, we additionally train 3D VAE-GAN.
Since this method trains one 3D-GAN for each object category, it is hard to apply the algorithm to 40 categories.
Therefore, we manually choose the classes for both training and testing, which have similar shapes to others: bench, chair and sofa.
Rest of the methods are evaluated on 40 classes.
The precision-recall curve of 3D shape reconstruction is depicted in Fig.~\ref{precision_recall}.

To verify the reconstruction results in various environments, we evaluate NOLBO on MS-COCO \cite{COCOdataset}, TUM \cite{TUMdataset} and KITTI dataset \cite{KITTI}.
Some of the 3D reconstructions of multi-object are displayed in Fig.~\ref{multiobject3Dreconstruction}.
To clearly show the reconstructed 3D shapes, objects are aligned in arbitrary viewpoint.
For more quantitative results of multi-object 3D reconstruction, see our supplementary.

\subsection{Multi-object Observation and SLAM}
To verify the probabilistic SLAM using our observation model, we choose KITTI dataset for autonomous driving.
Among 10 sequences of KITTI, 00, 05, 06 and 07 are used which have various loop closing spots.
To evaluate the aspect of data association, we compare our SLAM results with that of using latent variables from other object understanding networks.
Results using 128dim-version NOLBO is also compared.
Average precisions of the trajectory estimation are represented in Table.\ref{ATEandRPE}.
We also display some of the results in  Fig.~\ref{trj}.
For SLAM, any visual odometry can be used for initial trajectory estimation. However, to compare data association results more clearly, we use low-performance visual odometry based on simple visual feature matching procedure with known scale as shown in Fig.~\ref{trj}(b).
Results using our model with objectness estimated from the entire single scene show better loop closing performance than that of using the single-object model.
Even if our method is trained on indoor datasets, it can achieve the robust object observation in the wild and can consider the various loop closing points.
As shown in Table.\ref{ATEandRPE}, with high dimensional (128) latent variables, data association occasionally fails due to the bubble effect.
For more results, refer to see our supplementary.

\section{Conclusion}
We have proposed an observation model approximation for 3D multi-object in single 2D scene.
Since 3D objects in a shot are related to each other and follow a complex joint distribution, object-oriented probabilistic SLAM poses a challenge.
Therefore, we approximate the joint distribution of the multi-object present in a single 2D scene using VAE.
The jointly correlated multi-object observation is represented with factorized tractable distributions in the latent space.
Each observation can be estimated in consideration of the entire scene, by exploiting the existing multi-object detector structure.
Since our generative story involves the objectness, observation uncertainty naturally affects the probabilistic data association for SLAM.
The network is essentially an auto-encoder so that it can be used to estimate the 3D shape and pose of multi-object in a single shot.

\section*{Acknowledgement}
This work was supported by the National Research Foundation of Korea(NRF) grant funded by the Korea government(MSIP) (No. 2017R1A2B2002608), in part by Automation and Systems Research Institute (ASRI), and in part by the Brain Korea 21 Plus Project.

\onecolumn
\section{Appendix}
In this section, we present detailed formulations for probabilistic simultaneous localization and mapping (SLAM).
From Appendix I to III, we specify the probability distributions for our observation model, and introduce  formulations for several Bayesian inference methods:
in Appendix I, we introduce the distributions for latent variables in detail.
In Appendix II, we show how our approximated observation model can be adopted for probabilistic semantic SLAM, which can be performed using expectation-maximization (EM) method.
Examples of the maximum likelihood estimation (MLE) with our model is presented in Appendix III.


\section*{Appendix I : Lower Bound and Probability Distributions for Observation Model}
Combining Eqn. (4) and (6) of the paper, the lower bound $L$ of 
$p\left(
\mathcal{S}^f, \mathcal{L}, \mathcal{V}
\right)$
can be represented as the following:
\begin{align}
\nonumber
\log p\left(
\mathcal{S}^f, \mathcal{L}, \mathcal{V}
\right)
&\geq L
\\
\nonumber
&=
\sum_{k
}
-KL\left(
q_{\phi}\left(o_k|I\right) || p\left(o|\boldsymbol{l}_k\right)
\right)
\\
\nonumber
&
+
\sum_{k
}
\left(
-KL\left(
q_\phi\left(
\boldsymbol{z}^l_k | I
\right)
||
p_\psi\left(
\boldsymbol{z}^l_k | \boldsymbol{l}_k
\right)
\right)
+
\mathop{\mathbb{E}}_{q}
\left[
\log
p_\theta\left(
\boldsymbol{s}^{fo}_k | \boldsymbol{z}^l_k
\right)
\right]
\right)
\\
&
+
\sum_{k
}
\left(
-KL\left(
q_\phi\left(
\boldsymbol{z}^v_k | I
\right)
||
p\left(\boldsymbol{z}^v_k | \boldsymbol{v}_k\right)
\right)
+
\mathop{\mathbb{E}}_{q}
\left[
p\left(
\boldsymbol{R}^f_k | \boldsymbol{R}^z_k
\right)
\right]
\right)
.
\label{ELBO}
\end{align}
The probability $p\left(o|\boldsymbol{l}\right)$ representing the objectness according to the label is defined as:
\begin{align}
p\left(o|\boldsymbol{l}\right)
=
\begin{cases}
\epsilon &\text{, $\boldsymbol{l}\in\mathcal{L}^{\neg o}$, $o=$true}\\
1-\epsilon &\text{, $\boldsymbol{l}\in\mathcal{L}^{\neg o}$, $o=$false}\\
1-\epsilon^\prime &\text{, $\boldsymbol{l}\in\mathcal{L}^o$, \text{	} $o=$true}\\
\epsilon^\prime &\text{, $\boldsymbol{l}\in\mathcal{L}^o$, \text{	} $o=$false}
\end{cases}
\end{align}
where $\epsilon, \epsilon^\prime \ll 1$.
For convenience, we let $\epsilon=\epsilon^\prime$.
Here, $q_\phi\left(o | I \right)$ is assumed to be the Bernoulli distribution of objectness.

For the tractable prior distribution of $\boldsymbol{z}^l$, we assume $p_\psi\left(\boldsymbol{z}^l|\boldsymbol{l}\right) = \mathcal{N}\left(\boldsymbol{z}^l;\boldsymbol{\mu}\left(\boldsymbol{l}\right), \boldsymbol{I}\right)$.
In other words, we let $p_\psi\left(\boldsymbol{z}^l\right) = \sum_{\boldsymbol{l}} p\left(\boldsymbol{z}^l|\boldsymbol{l}\right) p\left(\boldsymbol{l}\right)$ be a gaussian mixture model (GMM).
$\boldsymbol{\mu}\left(\boldsymbol{l}\right) = f_\psi \left(\boldsymbol{l}\right)$ is obtained from prior network with parameter $\psi$, which is trained with VAE simultaneously.
To simplify the network structure, the variance is assumed to be $\boldsymbol{I}$.
Similarly, the variational likelihood $q_\phi$ is assumed to be a multivariate Gaussian; 
$q_{\phi} \left(\boldsymbol{z}^l|I\right) = \mathcal{N}(\boldsymbol{z}^l; \boldsymbol{\mu}^{\boldsymbol{s}l}, \left(\boldsymbol{\sigma}^{\boldsymbol{s}l}\right)^2 )$.

To estimate the orientation of objects, classifying the discretized angles can be trained easily as compared to the continuous regression \cite{craftCNNforviewpoint,3D-RCNN}.
For the probabilistic modeling and observation uncertainty, however, it is much useful to assume a noise model rather than a multinoulli distribution. 
Since $\boldsymbol{v}$ represented as radian is natural to follow a Gaussian distribution, we can let the latent variable be the angle directly by assuming $p\left(\boldsymbol{z}^v|\boldsymbol{v}\right) = \mathcal{N}\left(\boldsymbol{z}^v;\boldsymbol{v}, {\sigma}^2\boldsymbol{I}\right)$.
However, as the direct estimation of the angle is challenging to the network \cite{generatechair}, we let the trigonometric function values of $\boldsymbol{v}$ be the latent variables rather than $\boldsymbol{v}$ itself.
In other words, we let $p\left(\boldsymbol{z}^v|\boldsymbol{v}\right) = \mathcal{N}\left(\boldsymbol{z}^v; \left(\boldsymbol{\mu}^s,\boldsymbol{\mu}^c\right), \left(\boldsymbol{\sigma}^s, \boldsymbol{\sigma}^c\right)^2\boldsymbol{I}\right)$ be the prior of the latent variable related to the orientation \cite{directional,directional1}.
Assuming $\sin v$ and $\cos v$ are i.i.d., $\boldsymbol{\mu}^s$, $\boldsymbol{\mu}^c$, $\boldsymbol{\sigma}^s$ and $\boldsymbol{\sigma}^c$ can be represented as follows \cite{directional}:
\begin{align}
\nonumber
\mu^s_i,\mu^c_i &= e^{-\sigma^2/2} \left(\sin v_i,\cos v_i\right),
\\
\nonumber
\sigma^s_i &= 1/2-1/2\text{ } e^{-2\sigma^2}\cos2v_i-e^{-\sigma^2}\sin^2 v_i ,
\\
\sigma^c_i &= 1/2+1/2 \text{ } e^{-2\sigma^2}\cos2v_i-e^{-\sigma^2}\cos^2 v_i
,
\label{meanvarofwrappedgaussian}
\end{align}
where $v_i\in\boldsymbol{v}$.
We set $\sigma$ for $p\left(\boldsymbol{z}|\boldsymbol{v}\right)$ to $0.05\text{}rad \simeq 2.87^{\circ}$ \cite{ICRA2019}, because when the lower precision is used, learning easily diverges.
Similar to the prior, a variational likelihood for the orientation is assumed to be a wrapped normal distribution with trigonometric functions: $q_{\phi}\left(\boldsymbol{z}^v|\boldsymbol{s}^f\right) = \mathcal{N}\left(\boldsymbol{z}^v; \left(\boldsymbol{\mu}^{ss},\boldsymbol{\mu}^{sc}\right), \left(\boldsymbol{\sigma}^{ss}, \boldsymbol{\sigma}^{sc}\right)^2\boldsymbol{I}\right)$.
Therefore, for $q_{\phi}\left(\boldsymbol{z}^v|\boldsymbol{s}^f\right)$ we first infer $\sin \boldsymbol{v}^s,\cos \boldsymbol{v}^s$ and $\boldsymbol{\sigma}^s$ from the network, then calculate mean and variance similar to \eqref{meanvarofwrappedgaussian}.
In practice, we give an additional constraint $\sin^2 v + \cos^2 v = 1$ in order to consider that trigonometric functions are not independent.
Since the inferred latent variables are the trigonometric values of the pose, rotation matrix $\boldsymbol{R}^z$ related to $\boldsymbol{z}^v\sim q_\phi$ can directly be computed.

\section*{Appendix II : Object-oriented Probabilistic Semantic SLAM with Approximated Observation Model}

\subsection*{A. EM Formulation of Probabilistic Semantic SLAM}
Classical SLAM methods usually divide the problem into two parts: data associations in front-end, and pose optimization in back-end \cite{g2o,slam++,orbSLAM,categorySpecificSLAM}.
With these approaches, the huge error is inevitable when the false data association occurs in the front-end, since incorrectly determined data associations are hardly modified and thus bring a highly detrimental effect on pose estimation in the back-end.

To avoid this limitation, the complete SLAM formulation with probabilistic data association can be achieved using Expectation-Maximization (EM) method \cite{bowman2017probabilistic,IROS2018,ICRA2019}; both pose and data association can be optimized simultaneously.
Consider the localization and mapping problem with object-oriented features.
Our observer (camera or drone) collects a set of single shots $\{I_t\}^T_{t=1}$ as keyframes.
Let $\mathcal{X}=\{\mathscr{x}_t=\left(\boldsymbol{x}^{\mathscr{x}}_t, \boldsymbol{v}^{\mathscr{x}}_t\right)\}^T_{t=1}$ be the pose of an observer that navigates the unknown area;
$\boldsymbol{x}^{\mathscr{x}}_t$ and $\boldsymbol{v}^{\mathscr{x}}_t$ represent 3D position and orientation of the observer, respectively.
Assume we have a collection $\mathscr{L}=\{\ell_m = \left(\boldsymbol{l}_m, \boldsymbol{x}^{\ell}_m, \boldsymbol{v}^{\ell,global}_m\right)\}_{m=1}^M$ of $M$ landmarks; $\boldsymbol{l}$ stands for the category or instance-level label, and $\boldsymbol{x}^{\ell}_m$ and $\boldsymbol{v}^{\ell,global}_m$ denote 3D position and orientation of the landmark in the global coordinate, respectively.
Now suppose that the observer navigates around the area and obtain a set of 3D object observation $\mathcal{S}=\{\mathcal{S}^T_{t=1}\}$.
Here, $k$'th observation $\boldsymbol{s}_k = \left(\boldsymbol{s}^f_k,\boldsymbol{x}^{\boldsymbol{s}}_k\right) \in \mathcal{S}_t$ consists of the 3D position $\boldsymbol{x}^{\boldsymbol{s}}$ and the full shape $\boldsymbol{s}^f$ considering the object's orientation related to the observer.
The EM formulation for the probabilistic SLAM is expressed as follows \cite{bowman2017probabilistic,IROS2018}:
\begin{align}
w^t_{ij}
&=
\frac
{
	\sum_{\mathcal{D}_t^\prime \in \mathbb{D}_t\left(i,j\right)}
	p\left(
	\mathcal{S}_t|\mathcal{X},\mathscr{L},\mathcal{D}_t^\prime
	\right)
}
{
	\sum_{\mathcal{D}_t \in \mathbb{D}_t}
	p\left(
	\mathcal{S}_t|\mathcal{X},\mathscr{L},\mathcal{D}_t
	\right)	
}\text{ }\text{ }\text{ } \forall t,i,j
\label{EM_expectation}
\\
\mathcal{X}, \mathscr{L}
&=
\argmin_{\mathcal{X}, \mathscr{L}}
\sum_{t=1}^{T}
\sum_{k,j}
\sum_{\mathcal{D}_t^\prime \in \mathbb{D}_t\left(k,j\right)}
-w^t_{kj}
\log p\left(
\mathcal{S}_t|\mathcal{X},\mathscr{L},\mathcal{D}_t^\prime
\right).
\label{EM_maximization}
\end{align}
$\mathbb{D}_t$ is the set of all possible data associations 
$\mathcal{D}_t = \{\left(\alpha_k, \beta_k \right) \}^K_{k=1}$ representing that the object detection $\boldsymbol{s}_k$ of landmark $\ell_{\beta_k}$ was obtained from the observer state $\mathscr{x}_{\alpha_k}$.
Also,
$\mathbb{D}_t\left(i,j\right)\subseteq\mathbb{D}_t$ is the set of all possible data association $\mathcal{D}_t^\prime = \{\left(\alpha_k^\prime, \beta_k^\prime \right) \}$ such that $i$th detection is assigned to $j$th landmark.

Now suppose the data associations for $t$'th keyframe are fixed, then we have:
\begin{align}
p\left(
\mathcal{S}_t | \mathcal{X}, \mathscr{L}, \mathcal{D}_t
\right)
=
p\left(
\mathcal{S}_t | \mathcal{X}_\alpha, \mathscr{L}_\beta
\right),
\label{keyframe_obs}
\end{align}
where $\mathcal{X}_\alpha = \{\mathscr{x}_{\alpha_k}\}_{k=1}^K$ and $\mathscr{L}_\beta = \{\ell_{\beta_k}\}_{k=1}^K$.
Since we assume that $\mathcal{S}^f=\{\boldsymbol{s}^f_k\}_{k=1}^K$ is generated by $\mathcal{L} = \{\boldsymbol{l}_k\}_{k=1}^K$ and $\mathcal{V}=\{\boldsymbol{v}_k\}_{k=1}^K$, observation probability \eqref{keyframe_obs} can be split as follows:
\begin{align}
\nonumber
p\left(
\mathcal{S}_t | \mathcal{X}_\alpha, \mathscr{L}_\beta
\right)
&=
p\left(
\mathcal{S}^f_t, \boldsymbol{X}^{\boldsymbol{s}}_t
|
\boldsymbol{X}^{\mathscr{x}}_{\alpha}, \mathcal{V}^{\mathscr{x}}_{\alpha},
\mathcal{L}_{\beta}, \boldsymbol{X}^{\ell}_{\beta}, \mathcal{V}^{\ell,global}_{\beta}
\right)
\\
&=
p\left(
\boldsymbol{X}^{\boldsymbol{s}}_t | \boldsymbol{X}^{\mathscr{x}}_{\alpha},\boldsymbol{X}^{\ell}_{\beta}
\right)
p\left(
\mathcal{S}^f_t | \mathcal{L}_{\beta},\mathcal{V}^{\ell,global}_{\beta},\mathcal{V}^{\mathscr{x}}_{\alpha}
\right),
\label{factorizedobservationfork}
\end{align}
where $\boldsymbol{X}^{\boldsymbol{s}}_t=\{\boldsymbol{x}^s_k\}_{k=1}^K$,
$\boldsymbol{X}^{\mathscr{x}}_{\alpha} = \{\boldsymbol{x}^{\mathscr{x}}_{\alpha_k}\}_{k=1}^K$ and
$\boldsymbol{X}^{\ell}_{\beta} = \{\boldsymbol{x}^{\ell}_{\beta_k}\}_{k=1}^K$.
For the first factorized term in \eqref{factorizedobservationfork}, we can say that the set of feature's 3D position $\boldsymbol{X}^{\boldsymbol{s}}_t$ is determined by the set of observer's position $\boldsymbol{X}^{\mathscr{x}}_{\alpha}$ and the observed object's position $\boldsymbol{X}^{\ell}_{\beta}$.
The term
$
p\left(
\mathcal{S}^f_t | \mathcal{L}_{\beta},\mathcal{V}^{\ell,global}_{\beta},\mathcal{V}^{\mathscr{x}}_{\alpha}
\right)
$ in \eqref{factorizedobservationfork} denotes that the observer with states $\mathcal{V}^{\mathscr{x}}_{\alpha}$ observes 3D shapes of objects $\boldsymbol{S}^f_t$ with label $\mathcal{L}_{\beta}$ and pose $\mathcal{V}^{\ell,global}_{\beta}$.
It is equivalent that the observer observes the object $\mathcal{L}_{\beta}$ placed according to the local orientation $\mathcal{V}^{\ell,local}_{\alpha,\beta}$, which is related to the observer;
relation between $\boldsymbol{v}^{local}_{\alpha_k,\beta_k}\in\mathcal{V}^{\ell,local}_{\alpha,\beta}$ and $\boldsymbol{v}^{global}_{\beta_k}\in\mathcal{V}^{\ell,global}_{\beta}$ can be represented as:
\begin{align}
\boldsymbol{R}^{\boldsymbol{v}^{\ell,local}}_{\alpha_k,\beta_k} = \boldsymbol{R}^{\boldsymbol{v}^{\ell,global}}_{\beta_k}
\left(\boldsymbol{R}^{\boldsymbol{v}^{\mathscr{x}}}_{\alpha_k}\right)^{-1},
\end{align}
where $\boldsymbol{R}^{\boldsymbol{v}}$ denotes the rotation transform matrix according to the pose $\boldsymbol{v}$. Therefore, without loss of generality, we can rewrite \eqref{factorizedobservationfork} as the following:
\begin{align}
p\left(
\mathcal{S}_t | \mathcal{X}_\alpha, \mathscr{L}_\beta
\right)
=
p\left(
\boldsymbol{X}^{\boldsymbol{s}}_t | \boldsymbol{X}^{\mathscr{x}}_{\alpha},\boldsymbol{X}^{\ell}_{\beta}
\right)
p\left(
\mathcal{S}^f_t | \mathcal{L}_{\beta},\mathcal{V}^{\ell,local}_{\alpha, \beta}
\right)
\label{finalfactorizedobservationfork}
\end{align}
Substituting \eqref{finalfactorizedobservationfork} to \eqref{EM_expectation} and \eqref{EM_maximization} finally yields:
\begin{align}
w^t_{ij}
&=
\frac{
	\sum_{\mathcal{D}_t^\prime \in \mathbb{D}_t\left(i,j\right)}
	p\left(
	\boldsymbol{X}^{\boldsymbol{s}}_t | \boldsymbol{X}^{\mathscr{x}}_{\alpha^\prime},\boldsymbol{X}^{\ell}_{\beta^\prime}
	\right)
	p\left(
	\mathcal{S}^f_t | \mathcal{L}_{\beta^\prime},\mathcal{V}^{\ell,local}_{\alpha^\prime,\beta^\prime}
	\right)
}
{
	\sum_{\mathcal{D}_t \in \mathbb{D}_t}
	p\left(
	\boldsymbol{X}^{\boldsymbol{s}}_t | \boldsymbol{X}^{\mathscr{x}}_{\alpha},\boldsymbol{X}^{\ell}_{\beta}
	\right)
	p\left(
	\mathcal{S}^f_t | \mathcal{L}_{\beta},\mathcal{V}^{\ell,local}_{\alpha,\beta}
	\right)
}\text{ }\text{ }\text{ } \forall t,i,j
\label{EM_expectation_final}
\\
\mathcal{X}, \mathscr{L}
&=
\argmin_{\mathcal{X}, \mathscr{L}}
\sum_{t=1}^{T}
\sum_{k,j}
\sum_{\mathcal{D}_t^\prime \in \mathbb{D}_t\left(k,j\right)}
-w^t_{kj}
\log
p\left(
\boldsymbol{X}^{\boldsymbol{s}}_t | \boldsymbol{X}^{\mathscr{x}}_{\alpha},\boldsymbol{X}^{\ell}_{\beta}
\right)
p\left(
\mathcal{S}^f_t | \mathcal{L}_{\beta},\mathcal{V}^{\ell,local}_{\alpha, \beta}
\right).
\label{EM_maximization_final}
\end{align}

\subsection*{B. Variational Observation Model Approximation and EM for Probabilistic SLAM}
\subsubsection*{Expectation Step}
The evidence lower bound (ELBO) nearly reaches the target distribution when variational auto-encoder (VAE) is correctly converged.
Therefore, we can let the lower bound $L$ in \eqref{ELBO} approximately represent the joint distribution for the 3D shape observation probability
$
p\left(
\mathcal{S}^f | \mathcal{L},\mathcal{V}
\right)
$
in \eqref{finalfactorizedobservationfork} as follows:
\begin{align}
\nonumber
\log p\left(
\mathcal{S}^f, \mathcal{L}, \mathcal{V}
\right)
&=
\log p\left(\mathcal{S}^f | \mathcal{L}, \mathcal{V}\right)
+
\log p\left(\mathcal{V}\right)
+
\log p\left(\mathcal{L}\right)
\\
\nonumber
&\geq
c_0+c_1
\\
\nonumber
&+
\sum_{k
}
-KL\left(
q_{\phi}\left(o_k|I\right) || p\left(o|\boldsymbol{l}_k\right)
\right)
\\
\nonumber
&
+
\sum_{k
}
\left(
-KL\left(
q_\phi\left(
\boldsymbol{z}^l_k | I
\right)
||
p_\psi\left(
\boldsymbol{z}^l_k | \boldsymbol{l}_k
\right)
\right)
+
\mathop{\mathbb{E}}_{q}
\left[
\log
p_\theta\left(
\boldsymbol{s}^{fo}_k | \boldsymbol{z}^l_k
\right)
\right]
\right)
\\
&
+
\sum_{k
}
\left(
-KL\left(
q_\phi\left(
\boldsymbol{z}^v_k | I
\right)
||
p\left(\boldsymbol{z}^v_k | \boldsymbol{v}_k\right)
\right)
+
\mathop{\mathbb{E}}_{q}
\left[
p\left(
\boldsymbol{R}^f_k | \boldsymbol{R}^z_k
\right)
\right]
\right).
\end{align}
Take exponential to both sides, we have:
\begin{align}
p\left(\mathcal{S}^f | \mathcal{L}, \mathcal{V}\right)
\simeq
a
\prod_{k
}
\kappa_{KL}^o\left(\boldsymbol{l}_k;I\right)
\kappa_{KL}^z\left(\boldsymbol{l}_k;I\right)
\kappa_{KL}^z\left(\boldsymbol{v}_k;I\right)
\kappa_{E}\left(\boldsymbol{s}^{fo}_k\right)
\kappa_{E}\left(\boldsymbol{R}^{f}_k\right).
\label{approximatedlikelihood}
\end{align}
In \eqref{approximatedlikelihood}, $a$ is a constant, and
\begin{align}
\nonumber
\kappa_{KL}^o\left(
\boldsymbol{l}_k;I
\right)
&=
\exp\left(
-KL\left(
q_\phi\left(o_k|I\right)
||
p\left(o_k|\boldsymbol{l}_k\right)
\right)
\right)
\\
\nonumber
\kappa_{KL}^z\left(
\boldsymbol{l}_k;I
\right)
&=
\exp\left(
-KL\left(
q_\phi\left(\boldsymbol{z}^l_k|I\right)
||
p_\psi\left(\boldsymbol{z}^l_k|\boldsymbol{l}_k\right)
\right)
\right)
\\
\nonumber
\kappa_{KL}^z\left(
\boldsymbol{v}_k;I
\right)
&=
\exp\left(
-KL\left(
q_\phi\left(\boldsymbol{z}^v_k|I\right)
||
p\left(\boldsymbol{z}^v_k|\boldsymbol{v}_k\right)
\right)
\right)
\\
\nonumber
\kappa_{E}\left(\boldsymbol{s}^{fo}_k\right)
&=
\exp\left(
\mathop{\mathbb{E}}_{\boldsymbol{z}^l_k\sim q_\phi}
\left[
\log p_\theta\left(\boldsymbol{s}^{fo}_k|\boldsymbol{z}^l_k\right)
\right]
\right)
\\
\nonumber
\kappa_{E}\left(\boldsymbol{R}^{f}_k\right)
&=
\exp\left(
\mathop{\mathbb{E}}_{\boldsymbol{z}^v_k\sim q_\phi}
\left[
\log p\left(\boldsymbol{R}^f_k|\boldsymbol{R}^z_k\right)
\right]
\right).
\end{align}
Note that although the original joint probability for single scene is intractable and extremely challenging to express in factorized form,
we can have such factorized form as \eqref{approximatedlikelihood} that follow tractable Gaussians by leveraging the latent space.

For $\kappa_{KL}^o\left(\boldsymbol{l}_k;I\right)$, we have:
\begin{align}
\kappa_{KL}^o\left(\boldsymbol{l}_k;I\right)
\simeq
\begin{cases}
1, &\text{ $\boldsymbol{l}_k\in \mathcal{L}^o$,\text{ }\text{ } $q_{\phi^o}\left(o=\textit{true}|I\right)\simeq 1$}\text{ }\text{ }\text{ }\text{ }(a)
\\
0, &\text{ $\boldsymbol{l}_k\in \mathcal{L}^o$,\text{ }\text{ } $q_{\phi^o}\left(o=\textit{true}|I\right)\simeq 0$}\text{ }\text{ }\text{ }\text{ }(b)
\\
1, &\text{ $\boldsymbol{l}_k\in \mathcal{L}^{\neg o}$, $q_{\phi^o}\left(o=\textit{true}|I\right)\simeq 0$}\text{ }\text{ }\text{ }\text{ }(c)
\\
0, &\text{ $\boldsymbol{l}_k\in \mathcal{L}^{\neg o}$, $q_{\phi^o}\left(o=\textit{true}|I\right)\simeq 1$}\text{ }\text{ }\text{ }\text{ }(d)
\end{cases}
\label{kappaforobjectness}
\end{align}
Since our main concern is object-oriented SLAM, it is necessary to focus on the object observations.
To achieve this, we only calculate \eqref{kappaforobjectness} for the prior $p\left(o|\boldsymbol{l}\right)$ where $\boldsymbol{l}\in\mathcal{L}^o$.
Then the cases (a) and (b) of \eqref{kappaforobjectness} remain, which indicate that if the objectness $o$ of an observation is close to \textit{true}, \eqref{kappaforobjectness} goes to 1;
else 0.
In other words, \eqref{kappaforobjectness} can be seen as the probability of objectness when calculated only on $p\left(o|\boldsymbol{l}\right)$ for $\boldsymbol{l}\in\mathcal{L}^o$.
With this manner, all observations for $K$ regions can be used naturally by letting the probability of objectness affect to the weight calculation;
we can reflect both `more object-like' observations and `less-like' one into the weight.
There is no need to just abandon several observations which have the objectness under certain threshold, and give the equal weights to others.


Substituting \eqref{approximatedlikelihood} into \eqref{EM_expectation_final}, we have:
\begin{align}
w^t_{ij}
\simeq
\frac{
	\sum
	\prod
	p\left(
	\boldsymbol{x}^{\boldsymbol{s}}_k | \boldsymbol{x}^\mathscr{x}_{\alpha^\prime_k},\boldsymbol{x}^\ell_{\beta^\prime_k}
	\right)
	\kappa_{KL}^o\left( \boldsymbol{l}_{\beta^\prime_k};I_t\right)
	\kappa_{KL}^z\left( \boldsymbol{l}_{\beta^\prime_k};I_t\right)
	\kappa_{KL}^z\left( \boldsymbol{v}_{\alpha^\prime_k,\beta^\prime_k}^{\ell,local};I_t\right)
	\kappa_{E}\left(\boldsymbol{s}^{fo}_k\right)
	\kappa_{E}\left(\boldsymbol{R}^{f}_k\right)
}
{
	\sum
	\prod
	p\left(
	\boldsymbol{x}^{\boldsymbol{s}}_k | \boldsymbol{x}^\mathscr{x}_{\alpha_k},\boldsymbol{x}^\ell_{\beta_k}
	\right)
	\kappa_{KL}^o\left(\boldsymbol{l}_{\beta_k};I_t\right)
	\kappa_{KL}^z\left(\boldsymbol{l}_{\beta_k};I_t\right)
	\kappa_{KL}^z\left( \boldsymbol{v}_{\alpha_k,\beta_k}^{\ell,local};I_t\right)
	\kappa_{E}\left(\boldsymbol{s}^{fo}_k\right)
	\kappa_{E}\left(\boldsymbol{R}^{f}_k\right)
}
\label{approximatedweight}
\end{align}
We assume the 3D positions of $K$ observations are i.i.d.
Since
$\kappa_{E}\left(\boldsymbol{s}^{fo}\right)$ and
$\kappa_{E}\left(\boldsymbol{R}^{f}\right)$ are independent to $\mathcal{D}\in\mathbb{D}$, we can reduce the fraction.
Then \eqref{approximatedweight} can be expressed as:
\begin{align}
w^t_{ij}
=
\frac{
	\sum_{\mathcal{D}_t^\prime \in \mathbb{D}_t\left(i,j\right)}
	\prod_{k}
	p\left(
	\boldsymbol{x}^{\boldsymbol{s}}_k | \boldsymbol{x}^\mathscr{x}_{\alpha^\prime_k},\boldsymbol{x}^\ell_{\beta^\prime_k}
	\right)
	\kappa_{KL}^o\left( \boldsymbol{l}_{\beta^\prime_k};I_t\right)
	\kappa_{KL}^z\left( \boldsymbol{l}_{\beta^\prime_k};I_t\right)
	\kappa_{KL}^z\left( \boldsymbol{v}_{\alpha^\prime_k,\beta^\prime_k}^{\ell,local};I_t\right)
}
{
	\sum_{\mathcal{D}_t \in \mathbb{D}_t}
	\prod_{k}
	p\left(
	\boldsymbol{x}^{\boldsymbol{s}}_k | \boldsymbol{x}^\mathscr{x}_{\alpha_k},\boldsymbol{x}^\ell_{\beta_k}
	\right)
	\kappa_{KL}^o\left(\boldsymbol{l}_{\beta_k};I_t\right)
	\kappa_{KL}^z\left(\boldsymbol{l}_{\beta_k};I_t\right)
	\kappa_{KL}^z\left(\boldsymbol{v}_{\alpha_k,\beta_k}^{\ell,local};I_t\right)
}.
\label{reducedapproximatedweight}
\end{align}
Meanwhile, the KL-divergence term of $\kappa_{KL}^z\left(\boldsymbol{l}_{\beta_k};I_t\right)$ in \eqref{reducedapproximatedweight} is expanded as:
\begin{align}
KL\left(
q_\phi\left(\boldsymbol{z}|I_t\right)
||
p_\psi\left(\boldsymbol{z}|\boldsymbol{l}\right)
\right)
=
-\mathop{\mathbb{E}}_{\boldsymbol{z}^l\sim q_\phi}
\left[
\log p_\psi\left(\boldsymbol{z}^l|\boldsymbol{l}\right)
\right]
-
H\left(
q_\phi\left(\boldsymbol{z}^l|I_t\right)
\right).
\end{align}
The prior of $\boldsymbol{z}^l$ is assumed to be a multivariate Gaussian: $p_\psi\left(\boldsymbol{z}^l|\boldsymbol{l}\right) = \mathcal{N}\left(\boldsymbol{z}^l;\boldsymbol{\mu}\left(\boldsymbol{l}\right), \boldsymbol{I}\right)$.
The variational likelihood is also represented as $q_{\phi} \left(\boldsymbol{z}^l_k|I\right) = \mathcal{N}(\boldsymbol{z}; \boldsymbol{\mu}^{\boldsymbol{s}l}_k, \left(\boldsymbol{\sigma}^{\boldsymbol{s}l}_k\right)^2 \boldsymbol{I})$.
Note that $\boldsymbol{\mu}^{\boldsymbol{s}l}_k$ and $\boldsymbol{\sigma}^{\boldsymbol{s}l}_k$ are the variables estimated from $I$, using the encoder.
In other words, these variables are encoded features from the observed $I$.
With the prior and the variational likelihood, we have:
\begin{align}
\nonumber
KL&\left(
q_\phi\left(\boldsymbol{z}_k^l|I_t\right)
||
p_\psi\left(\boldsymbol{z}^l|\boldsymbol{l}\right)
\right)
\\
\nonumber
&=
-\mathop{\mathbb{E}}_{\boldsymbol{z}^l_k\sim q_{\phi}}
\left[
-\log Z 
- \frac{1}{2} \lVert \boldsymbol{z}^l - \boldsymbol{\mu}\left(\boldsymbol{l}\right) \rVert^2_{}
\right]
-
H\left(
q_{\phi}\left(\boldsymbol{z}^l_k|I_t\right)
\right)
\\
&=
\log Z + \frac{1}{2} \lVert \boldsymbol{\mu}^{\boldsymbol{s}l}_k - \boldsymbol{\mu}\left(\boldsymbol{l}\right) \rVert^2_{}
+ \frac{1}{2} tr \left(\left(\boldsymbol{\sigma}^{\boldsymbol{s}l}_k\right)^2 \boldsymbol{I} \right)
-
H\left(
q_{\phi}\left(\boldsymbol{z}_k^l|I_t\right)
\right)
\label{KLexpandforlabel}
\end{align}
where $Z$ is the normalization constant. With \eqref{KLexpandforlabel}, we can express $\kappa_{KL}^z\left(\boldsymbol{l}_{\beta_k};I_t\right)$ as:
\begin{align}
\nonumber
\kappa_{KL}^z\left(\boldsymbol{l}_{\beta_k};I_t\right)
&=
\exp \left(
-KL\left(
q_\phi\left(\boldsymbol{z}^l_k|I_t\right)
||
p_\psi\left(\boldsymbol{z}^l|\boldsymbol{l}_{\beta_k}\right)
\right)
\right)
\\
\nonumber
&=
\frac{1}{Z}\exp \left(
-
\frac{1}{2}
\lVert
\boldsymbol{\mu}^{\boldsymbol{s}l}_k - \boldsymbol{\mu}\left(\boldsymbol{l}_{\beta_k}\right)
\rVert^2_{}
\right)
\frac{\exp\left(H\left(
	q_{\phi}\left(\boldsymbol{z}^l_k|I_t\right)
	\right)\right)}
{\frac{1}{2} tr \left(\left(\boldsymbol{\sigma}^{\boldsymbol{s}l}\right)^2 \boldsymbol{I} \right)}
\\
&=
p_\psi \left( \boldsymbol{z} = \boldsymbol{\mu}^{\boldsymbol{s}l}_k | \boldsymbol{l}_{\beta_k} \right)
\frac{\exp\left(H\left(
	q_{\phi}\left(\boldsymbol{z}^l|I_t\right)
	\right)\right)}
{\exp\left(
	\frac{1}{2} tr \left(\left(\boldsymbol{\sigma}^{\boldsymbol{s}l}\right)^2 \boldsymbol{I} \right)
	\right)
}.
\label{kappaforlabel}
\end{align}
Meanwhile, we have assumed that $\boldsymbol{z}$ for angles follows Gaussian $p\left(\boldsymbol{z};\boldsymbol{v},\sigma^2\boldsymbol{I}\right)$ and the network is trained with trigonometric values of $\boldsymbol{z}$. The encoder thus infers the trigonometric values of $\boldsymbol{z}$ and precision: $\sin \boldsymbol{v}^{\boldsymbol{s}}$, $\cos \boldsymbol{v}^{\boldsymbol{s}}$ and $\boldsymbol{\sigma}^{\boldsymbol{s}}$. Using these values, we can calculate $\boldsymbol{v}^{\boldsymbol{s}}$. 
Therefore, similar to \eqref{kappaforlabel}, $\kappa_{KL}^z\left(\boldsymbol{v}^{\ell,local}_{\beta_k};I_t\right)$ in \eqref{reducedapproximatedweight} can be represented as follows:
\begin{align}
\kappa_{KL}^z\left(\boldsymbol{v}^{\ell,local}_{\beta_k};I\right)
=
p \left( \boldsymbol{z} = \boldsymbol{v}^{\boldsymbol{s}}_k | \boldsymbol{v}^{\ell,local}_{\beta_k}\right)
\frac{\exp\left(H\left(
	q_{\phi}\left(\boldsymbol{z}^v_k|I_t\right)
	\right)\right)}
{\exp\left(
	\frac{1}{2} tr \left(\left(\boldsymbol{\sigma}^{\boldsymbol{s}}_k\right)^2 \boldsymbol{I} \right)
	\right)
}.
\label{kappafororientation}
\end{align}
Since the exponential terms in \eqref{kappaforlabel} and \eqref{kappafororientation} are independent not to the data association but to $I$, substituting \eqref{kappaforlabel} and \eqref{kappafororientation} into \eqref{reducedapproximatedweight} and reducing the fraction finally yields:
\begin{align}
w^t_{ij}
\simeq
\frac{
	\sum_{\mathcal{D}_t^\prime \in \mathbb{D}_t\left(i,j\right)}
	\prod_{k}
	p\left(
	\boldsymbol{x}^{\boldsymbol{s}}_k | \boldsymbol{x}^\mathscr{x}_{\alpha^\prime_k},\boldsymbol{x}^\ell_{\beta^\prime_k}
	\right)
	\kappa_{KL}^o\left(\boldsymbol{l}_{\beta^\prime_k};I_t\right)
	p_\psi \left( \boldsymbol{z} = \boldsymbol{\mu}^{\boldsymbol{s}l}_k | \boldsymbol{l}_{\beta^\prime_k}\right)
	p \left( \boldsymbol{z} = \boldsymbol{v}^{\boldsymbol{s}}_k  | \boldsymbol{v}^{\ell,local}_{\alpha^\prime_k,\beta^\prime_k}\right)
}
{
	\sum_{\mathcal{D}_t \in \mathbb{D}_t}
	\prod_{k}
	p\left(
	\boldsymbol{x}^{\boldsymbol{s}}_k | \boldsymbol{x}^\mathscr{x}_{\alpha_k},\boldsymbol{x}^\ell_{\beta_k}
	\right)
	\kappa_{KL}^o\left(\boldsymbol{l}_{\beta_k};I_t\right)
	p_\psi \left( \boldsymbol{z} = \boldsymbol{\mu}^{\boldsymbol{s}l}_k | \boldsymbol{l}_{\beta_k}\right)
	p \left( \boldsymbol{z} = \boldsymbol{v}^{\boldsymbol{s}}_k  | \boldsymbol{v}^{\ell,local}_{\alpha_k,\beta_k}\right)
}
\label{weightfinal}
\end{align}
Note that \eqref{weightfinal} can be seen as the result of replacing the original observation model
$
p\left(
\mathcal{S}^f | \mathcal{L},\mathcal{V}^{\ell,local}
\right)
$
in \eqref{EM_expectation_final} with
$
\kappa_{KL}^o\left(\boldsymbol{l};I\right)
$
,
$
p_\psi \left( \boldsymbol{z}^l = \boldsymbol{\mu}^{\boldsymbol{s}l}_k | \boldsymbol{l}\right)
$
and
$
p \left( \boldsymbol{z}^v = \boldsymbol{v}^{\boldsymbol{s}}_k | \boldsymbol{v}^{\ell,local}\right)
$
, which are objectness and the priors of latent variables respectively.
Therefore, when the objectness $o_k$, shape feature $\boldsymbol{\mu}^{\boldsymbol{s}l}_k$ and orientation feature $\boldsymbol{v}^{\boldsymbol{s}}_k$ are obtained from the encoder of VAE, we can directly use these features as observations of 3D shape to have approximated weights for EM.

\subsubsection*{Maximization Step}
In the same manner as the expectation step, we can further exploit the approximated observation model in \eqref{approximatedlikelihood} for the maximization step.
Note that $\kappa_{E}$ terms in \eqref{approximatedlikelihood} are independent to $\mathcal{X}$ and $\mathcal{L}$.
Substituting \eqref{approximatedlikelihood} into \eqref{EM_maximization_final}, we have:
\begin{align}
\mathcal{X}, \mathscr{L}
=
\argmin_{\mathcal{X}, \mathscr{L}}
\sum_{t,k,j}
-w^t_{kj}
\log
p\left(
\boldsymbol{x}^{\boldsymbol{s}}_k | \boldsymbol{x}^\mathscr{x}_t,\boldsymbol{x}^\ell_j
\right)
\kappa_{KL}^z\left(
\boldsymbol{l}_{j} ;I
\right)
\kappa_{KL}^z\left(
\boldsymbol{v}^{\ell,local}_{j} ;I
\right)
.
\label{EM_maximization_1}
\end{align}
With \eqref{kappaforlabel} and \eqref{kappafororientation}, we get:
\begin{align}
\mathcal{X}, \mathscr{L}
=
\argmin_{\mathcal{X}, \mathscr{L}}
\sum_{t,k,j}
-w^t_{kj}
\log
p\left(
\boldsymbol{x}^{\boldsymbol{s}}_k | \boldsymbol{x}^\mathscr{x}_t,\boldsymbol{x}^\ell_j
\right)
p_\psi \left( \boldsymbol{z} = \boldsymbol{\mu}^{\boldsymbol{s}l}_k | \boldsymbol{l}_{j}\right)
p \left( \boldsymbol{z} = \boldsymbol{v}^{\boldsymbol{s}}_k  | \boldsymbol{v}^{\ell,local}_{t,j}\right)
.
\label{EM_maximization_2}
\end{align}
Since $\mathscr{L}=\{\ell_m = \left(\boldsymbol{l}_m, \boldsymbol{x}^\ell_m, \boldsymbol{v}^{\ell,global}_m\right)\}_{m=1}^M$ and $\boldsymbol{v}^{\ell,global}$ is represented with $\boldsymbol{v}^{\ell,local}$ and $\boldsymbol{v}^\mathscr{x}$, we can split \eqref{EM_maximization_2} as:
\begin{align}
\mathcal{X}, \mathcal{X}^\ell
&=
\argmin_{\mathcal{X}, \mathcal{X}^\ell}
\sum_{t,k,j}
-w^t_{kj}
\log
p\left(
\boldsymbol{x}^{\boldsymbol{s}}_k | \boldsymbol{x}^\mathscr{x}_t,\boldsymbol{x}^\ell_j
\right)
p \left( \boldsymbol{z} = \boldsymbol{v}^{\boldsymbol{s}}_k  | \boldsymbol{v}^{local}_{t,j}\right),
\label{EM_maximization_frac_x}
\\
\mathcal{L}^M
&=
\argmin_{\mathcal{L}^M}
\sum_{t,k,j}
-w^t_{kj}
\log
p_\psi \left( \boldsymbol{z} = \boldsymbol{\mu}^{\boldsymbol{s}l}_k | \boldsymbol{l}_{j}\right)
,
\label{EM_maximization_frac_label_appendix}
\end{align}
where $\mathcal{X}^\ell = \{\boldsymbol{x}^\ell_m, \boldsymbol{v}^{\ell,global}_m\}_{m=1}^M$ and $\mathcal{L}^M=\{\boldsymbol{l}_m\}_{m=1}^M$.
Similar to the expectation step, the maximization step can be calculated using the latent features $\boldsymbol{\mu}^{\boldsymbol{s}l}$ and $\boldsymbol{v}^{\boldsymbol{s}}$ encoded from the encoder.
Therefore, instead of using the complex 3D shape $\boldsymbol{s}^f$ itself or the intractable single view scene $I$ directly, we can perform EM for probabilistic SLAM only with the encoded latent features and their priors.
Since the conditional priors $p_\psi\left(\boldsymbol{z}^l|\boldsymbol{l}\right)$ and $p\left(\boldsymbol{z}^v|\boldsymbol{v}\right)$ are Gaussians, we can numerically calculate the optimal solutions of \eqref{EM_maximization_frac_x} and \eqref{EM_maximization_frac_label_appendix}.

Pose and angle are the continuous values, and various optimization techniques can be applied to obtain solutions of \eqref{EM_maximization_frac_x}. However, for \eqref{EM_maximization_frac_label_appendix}, we have no choice but to substitute all the possible labels to obtain the optimal solutions, because we assume $\mathcal{L}$ to be a set of the finite number of labels.
Meanwhile, since $p_\psi\left(\boldsymbol{z}^l|\boldsymbol{l}\right) = \mathcal{N}\left(\boldsymbol{z}; \boldsymbol{\mu}^l, \boldsymbol{I}\right)$ and $\boldsymbol{\mu}\left(\boldsymbol{l}\right)=f_\psi\left(\boldsymbol{l}\right)$,  we can say that the label $\boldsymbol{l}$ is projected to $\boldsymbol{\mu}\left(\boldsymbol{l}\right)$ with nonlinear regression function $f_\psi$.
In this point of view, we can redefine landmark as $\ell = \left(\boldsymbol{\mu}\left(\boldsymbol{l}\right), \boldsymbol{x}^{\ell}, \boldsymbol{v}^{\ell,global}\right)$.
Therefore, substituting $\boldsymbol{l}$ is the same as substituting $\boldsymbol{\mu}\left(\boldsymbol{l}\right)$ in the end, and we can exploit the continuous nonlinear regression of label $\boldsymbol{l}$ similar to \cite{ICRA2019}; we now concentrate on $\boldsymbol{\mu}\left(\boldsymbol{l}\right)$ rather than $\boldsymbol{l}$.
For simplicity, let $\boldsymbol{\mu}\left(\boldsymbol{l}\right) = \boldsymbol{\mu}^l$.
Then \eqref{EM_maximization_frac_label_appendix} can be expressed as:
\begin{align}
{\mathcal{U}}
&=
\argmin_{\mathcal{U}}
\sum_{t,k,j}
-w^t_{kj}
\log
p_\psi \left( \boldsymbol{z} = \boldsymbol{\mu}^{\boldsymbol{s}l}_k | \boldsymbol{\mu}_{j}^l\right)
,
\label{EM_maximization_frac_mu}
\end{align}
where $\mathcal{U}=\{\boldsymbol{\mu}_j^l\}$. As \eqref{EM_maximization_frac_mu} is sum of logarithmic Gaussians, we can differentiate \eqref{EM_maximization_frac_mu} with respect to $\boldsymbol{\mu}^l$ and obtain optimal solutions. The optimal $\boldsymbol{\mu}_j^l$ is then represented as the following:
\begin{align}
{\boldsymbol{\mu}_j^l}
=
\sum_{t,k}-w^t_{kj}\boldsymbol{\mu}^{\boldsymbol{s}l}_k
.
\label{optimal_mu}
\end{align}
By \eqref{optimal_mu}, we can numerically calculate the optimal feature vector $\boldsymbol{\mu}^l$, rather than selecting the optimal label for the shape.
It can be seen as updating shape features defined in continuous latent space, and shape estimation from multiple observation is achievable.
Since the finite label set is no longer needed, assumption of the static landmarks is also unnecessary.
Consequently, each observation can be regarded as an independent landmark, and we can calculate the similarities between them as \eqref{weightfinal} and then perform optimization with \eqref{EM_maximization_frac_x} and \eqref{optimal_mu} subsequently.

As the EM formulations are finally modified as \eqref{EM_maximization_frac_x} and \eqref{optimal_mu}, latent variables of other methods can be adopted for SLAM such as vanilla 2D-3D auto-encoder (AE), vanilla VAE (vVAE) \cite{vae} and TLNet \cite{TLNet}. We can extract latent variables of these algorithms with region of interest (RoI) image obtained by any other multi-object detector, and perform SLAM optimizations. 

\section*{Appendix III : MLE with Approximated Observation Model}
The approximated observation model can be used for classifying the categories or instances, or estimating the poses of the observed objects by Maximum Likelihood Estimation (MLE).
The classification problem for the label set $\mathcal{L}$ of multi-object in $I$ is given as:
\begin{align}
{\mathcal{L}}
=
\argmax_{\mathcal{L}}
p\left(
\mathcal{S}^f | \mathcal{L}
\right).
\label{MLEproblemclassification}
\end{align}
Using \eqref{approximatedlikelihood}, we have:
\begin{align}
\nonumber
p\left(
\mathcal{S}^f | \mathcal{L}
\right)
&=
a
\int_{\mathcal{V}}
p\left(
\mathcal{S}^f|\mathcal{L},\mathcal{V}
\right)
\mathop{\underline{
		p\left(
		\mathcal{V}	
		\right)
}}_{const}
d \mathcal{V}
\\
\nonumber
&\simeq
a^\prime
\int_{\mathcal{V}}
\prod_{k}
\kappa_{KL}^z\left(\boldsymbol{l}_k;I\right)
\kappa_{KL}^z\left(\boldsymbol{v}_k;I\right)
\kappa_{E}\left(\boldsymbol{s}^{fo}_k\right)
\kappa_{E}\left(\boldsymbol{R}^{v}_k\right)
d\mathcal{V}
\\
&=
a^\prime
\prod_{k}
\kappa_{KL}^z\left(\boldsymbol{l}_k;I\right)
\kappa_{E}\left(\boldsymbol{s}^{fo}_k\right)
\kappa_{E}\left(\boldsymbol{R}^{v}_k\right)
\int_{\mathcal{V}}
\kappa_{KL}^z\left(\boldsymbol{v}_k;I\right)
d\mathcal{V}.
\label{integralwithapproximatedobservationmodel}
\end{align}
Note that only the term
$\kappa_{KL}^z\left(\boldsymbol{l};I\right)$ is related to the label set $\mathcal{L}$. Substituting \eqref{integralwithapproximatedobservationmodel} into \eqref{MLEproblemclassification}, we have:
\begin{align}
{\mathcal{L}}
\simeq
\argmax_{\mathcal{L}}
\prod_{k}
\kappa_{KL}^z\left(\boldsymbol{l}_k;I\right).
\label{MLEproblemwithkappa}
\end{align}
Putting \eqref{kappaforlabel} into \eqref{MLEproblemwithkappa} finally yields:
\begin{align}
\nonumber
{\mathcal{L}}
&\simeq
\argmax_{\mathcal{L}}
\prod_{k}
p_\psi \left( \boldsymbol{z} = \boldsymbol{\mu}^{\boldsymbol{s}l}_k | \boldsymbol{l}_k \right)
\frac{\exp\left(H\left(
	q_{\phi}\left(\boldsymbol{z}^l_k|I\right)
	\right)\right)}
{\exp\left(
	\frac{1}{2} tr \left(\left(\boldsymbol{\sigma}^{\boldsymbol{s}l}_k\right)^2 \boldsymbol{I} \right)
	\right)
}
\\
&=
\argmax_{\mathcal{L}}
\prod_{k}
p_\psi \left( \boldsymbol{z} = \boldsymbol{\mu}^{\boldsymbol{s}l}_k | \boldsymbol{l}_k \right)
\\
&=
\argmax_{\boldsymbol{l_1} }
p_\psi \left( \boldsymbol{z} = \boldsymbol{\mu}^{\boldsymbol{s}l}_1 | \boldsymbol{l}_1 \right)
, ...,
\argmax_{\boldsymbol{l_K}}
p_\psi \left( \boldsymbol{z} = \boldsymbol{\mu}^{\boldsymbol{s}l}_K | \boldsymbol{l}_K \right)
\label{MLEproblemfinal}
\end{align}
Since $\mathcal{L}=\{\boldsymbol{l}_k\}$, we can simply have:
\begin{align}
{\boldsymbol{l}}
\simeq
\argmax_{\boldsymbol{l}}
p_\psi \left( \boldsymbol{z} = \boldsymbol{\mu}^{\boldsymbol{s}l} | \boldsymbol{l} \right).
\end{align}
Similar to the classification, the approximated solution of the object pose estimation problem can be achieved by using \eqref{approximatedlikelihood} and \eqref{kappafororientation} as the following:
\begin{align}
\nonumber
{\boldsymbol{v}}
&=
\argmax_{\boldsymbol{v}}
p\left(
\boldsymbol{s}^f | \boldsymbol{v}	
\right)
\\
&\simeq
\argmax_{\boldsymbol{v}}
p \left(
\boldsymbol{z} = \boldsymbol{v}^{\boldsymbol{s}} | \boldsymbol{v}
\right).
\label{MLEproblempose}
\end{align}

\twocolumn

{\small
\bibliographystyle{ieee}
\bibliography{egbib}}

\end{document}